\def\year{2021}\relax
\documentclass[letterpaper]{article} 
\usepackage{aaai21}  
\usepackage{times}  
\usepackage{helvet} 
\usepackage{courier}  
\usepackage[hyphens]{url}  
\usepackage{graphicx} 
\usepackage{graphicx}  
\usepackage{natbib}  
\usepackage{caption} 
\usepackage{resizegather}
\usepackage{mathrsfs}  
\usepackage{mathtools}
\usepackage{amssymb}
\usepackage[switch]{lineno}
\usepackage{bm}
\def\va{{\bm{a}}}

\def\vu{{\bm{u}}}
\def\vv{{\bm{v}}}
\def\vw{{\bm{w}}}
\def\vx{{\bm{x}}}
\def\vy{{\bm{y}}}
\def\vz{{\bm{z}}}

\def\mA{{\bm{A}}}

\def\mU{{\bm{U}}}
\def\mV{{\bm{V}}}
\def\mW{{\bm{W}}}
\def\mX{{\bm{X}}}
\def\mY{{\bm{Y}}}

\title{Why Adversarial Interaction Creates Non-Homogeneous Patterns: A Pseudo-Reaction-Diffusion Model for Turing Instability}
\author {
	Litu Rout\\
}

\affiliations{
	Space Applications Centre \\
	Indian Space Research Organisation\\
	lr@sac.isro.gov.in
}

\begin{document}
	\maketitle
	
	\begin{abstract}
		Long after Turing's seminal Reaction-Diffusion (RD) model, the elegance of his fundamental equations alleviated much of the skepticism surrounding pattern formation. Though Turing model is a simplification and an idealization, it is one of the best-known theoretical models to explain patterns as a reminiscent of those observed in nature. Over the years, concerted efforts have been made to align theoretical models to explain patterns in real systems. The apparent difficulty in identifying the specific dynamics of the RD system makes the problem particularly challenging. Interestingly, we observe Turing-like patterns in a system of neurons with adversarial interaction. In this study, we establish the involvement of Turing instability to create such patterns. By theoretical and empirical studies, we present a \textit{pseudo-reaction-diffusion} model to explain the mechanism that may underlie these phenomena. While supervised learning attains homogeneous equilibrium, this paper suggests that the introduction of an adversary helps break this homogeneity to create non-homogeneous patterns at equilibrium. Further, we prove that randomly initialized gradient descent with over-parameterization can converge exponentially fast to an $\epsilon$-stationary point even under adversarial interaction. In addition, different from sole supervision, we show that the solutions obtained under adversarial interaction are not limited to a tiny subspace around initialization.
	\end{abstract}
	
	\section{Introduction}
	In this paper, we intend to demystify an interesting phenomenon: adversarial interaction between generator and discriminator creates non-homogeneous equilibrium by inducing Turing instability in a Pseudo-Reaction-Diffusion (PRD) model. This is in contrast to supervised learning where the identical model achieves homogeneous equilibrium while maintaining spatial symmetry over iterations. 
	
	Recent success of Generative Adversarial Networks (GANs)~\cite{goodfellow2014generative,arjovsky2017wasserstein} has led to exciting applications in a wide variety of tasks~\cite{luc2016semantic,zhu2017unpaired,ledig2017photo,engin2018cycle,rout2020s2a}. In adversarial learning paradigm, it is often required that a particular sample is generated subject to a conditional input. Typically, conditional GANs are employed to meet these demands~\cite{mirza2014conditional}. Further, it has been reported in copious literature that supervised learning with adversarial regularization performs better than sole supervision~\cite{ledig2017photo,rout2020alert,wang2018esrgan,wang2016generative,karacan2016learning,sarmad2019rl}.	In all these prior works, one may notice several crucial properties of adversarial interaction. It is worth emphasizing that adversarial learning owes its benefits to the continuously evolving loss function which otherwise is extremely difficult to model. Motivated by these findings, we uncover another interesting property of adversarial training. We observe that adversarial interaction helps break the spatial symmetry and homogeneity to create non-homogeneous patterns in weight space.
	
	The reason for studying these phenomena is multi-fold. The fact that adversarial interaction exhibits Turing-like patterns creates a dire need to investigate its connections to nature. In particular, these patterns often emerge in real world systems, such as butterfly wings, zebra, giraffe and leopard~\cite{turing1952chemical,meinhardt1982models,rauch2004role, nakamasu2009interactions,kondo2010reaction}. Interestingly, adversarial training captures some intricacies of this complex biological process that create evolutionary patterns in neural networks. Furthermore, it is important to understand neural synchronization in human brain to design better architectures~\cite{budzynski2009introduction}. This paper is intended to shed light on some of these aspects. 
	
	It is widely believed that fully connected networks already capture certain  important properties of deep learning~\cite{saxe2014exact,li2018learning}. While one may wish to extend these analyses to more complex networks, it may not allow a comprehensive study of various fundamental aspects in the nascent state of understanding. Besides, the complexity involved in studying the Reaction-Diffusion (RD) dynamics of a large neural network is enormous. For this reason, we study two layer neural networks and focus more on the theory of Turing-like patterns.

	While dynamical systems governed by different equations exhibit different patterns, it is crucial to study the dynamics through \textit{reaction and diffusion} terms that laid the foundation of pattern formation~\cite{turing1952chemical}. Our key observation:
	\begin{center}
		\textit{A system in which a generator and a discriminator adversarially interact with each other exhibits Turing-like patterns in the hidden layer and top layer of a two layer generator network with ReLU activation.}
	\end{center}
	To provide a thorough explanation to these empirical findings, we derive the governing dynamics of a PRD model. 
	
	From another perspective, the generator provides a short-range positive feedback as it tries to minimize the empirical risk directly. On the other hand, the discriminator provides a long-range negative feedback as it tries to maximize the generator cost. Since the adversary discriminates between real and fake samples, it indirectly optimizes the primary objective function. It is safe to assume that such signals from the discriminator to the generator form the basis of long-range negative feedback as studied by~\citeauthor{rauch2004role}.  
	
	
	\section{Preliminaries}
	\label{prelim}
	\subsubsection{Notations}
	Bold upper-case letter $\mA$ denotes a matrix. Bold lower-case letter $\va$ denotes a vector. Normal lower-case letter $a$ denotes a scalar. $\left \|. \right \|_2$ represents Euclidean norm of a vector and spectral norm of a matrix. $\left \|. \right \|_F$ represents Frobenius norm of a matrix. $\lambda_{\min}(.)$ and $\lambda_{\max}(.)$ denote smallest and largest eigen value of a matrix. $d x$ represents derivative of $x$ and $\partial x$ represents its partial derivative. For $g: \mathbb{R}^{d} \rightarrow \mathbb{R}$,  $\nabla g$ and $\nabla^2 g$  denote gradient and Laplacian of $g$, respectively. $[m]$ denotes the set $\left \{1,2,\dots,m \right\}$.
	
	\subsubsection{Problem Setup}
	\label{ps}
	Consider that we are given $n$ training samples $\left \{\left ( \vx_p, \vy_p \right )  \right \}_{p=1}^{n} \subset \mathbb{R}^{d_{in}} \times \mathbb{R}^{d_{out}}$. Formally, we use the following notations to represent two layer neural networks with rectified linear unit (ReLU) activation function $(\sigma(.))$. 
	\begin{equation}
	f\left ( \mU,\mV,\vx \right ) = \frac{1}{\sqrt{d_{out} m}}\mV\sigma\left (\mU \vx \right)
	\end{equation}
	Here, $\mU \in \mathbb{R}^{m \times d_{in}}$ and $\mV \in \mathbb{R}^{d_{out} \times m}$. Let us denote $\vu_j = \mU_{j,:}$ and $\vv_j = \mV_{:,j}$. The scaling factor $\frac{1}{\sqrt{d_{out} m}}$ is derived from Xavier initialization~\cite{glorot2010understanding}. In supervised learning, the training is carried out by minimizing the $l_2$ loss over data as given by 
	\begin{equation}
	\begin{split}
	\mathcal{L}_{sup}\left ( \mU, \mV \right ) &= \frac{1}{2} \sum_{p=1}^{n} \left \| \frac{1}{\sqrt{d_{out} m}}\mV\sigma\left (\mU \vx_p \right) - \vy_p \right \|_2^2 \\
	&=\frac{1}{2} \left \| \frac{1}{\sqrt{d_{out} m}}\mV\sigma\left (\mU \mX \right) - \mY \right \|_{F}^2.
	\end{split}
	\end{equation}
	The input data points are represented by $\mX = \left ( \vx_1, \vx_2, \dots, \vx_n \right ) \in \mathbb{R}^{d_{in}\times n}$ and corresponding labels by $\mY = \left ( \vy_1, \vy_2, \dots, \vy_n \right ) \in \mathbb{R}^{d_{out}\times n}$. In regularized adversarial learning, the generator cost is augmented with an adversary:
	\begin{equation}
	\begin{split}
	\mathcal{L}_{aug}\left ( \mU, \mV, \mW, \va \right ) & = \underset{\mathcal{L}_{sup}}{\underbrace{\frac{1}{2} \left \| \frac{1}{\sqrt{d_{out} m}}\mV\sigma\left (\mU \mX \right) - \mY \right \|_{F}^2}} \\ &-\underset{\mathcal{L}_{adv}}{\underbrace{\frac{1}{m\sqrt{d_{out}}} \sum_{p=1}^{n}\va^T \sigma\left ( \mW \mV\sigma\left (\mU \vx_p \right)\right)}}.
	\end{split}
	\end{equation}
	The adversary, $g\left ( \mW,\va, y \right ) = \frac{1}{\sqrt{m}}\va^T \sigma\left (\mW \vy  \right ): \mathbb{R}^{d_{out}}\rightarrow \mathbb{R}$ is a two layer network with ReLU activation. Here, $\mW \in \mathbb{R}^{m\times d_{out}}$ and $\va \in \mathbb{R}^{m}$. The discriminator cost is exactly identical to the critic of WGAN with gradient penalty~\cite{gulrajani2017improved}. We follow the common practice to train generator and discriminator alternatively using Wasserstein distance. In this study, $\mathcal{L}_{aug}$ is considered as the equivalent of a continuous field in a RD system~\cite{turing1952chemical}.
	
	\subsubsection{Learning Algorithm}
	We consider vanilla gradient descent with random initialization as our learning algorithm to minimize both supervised and augmented objective. For instance, we update each trainable parameter in augmented objective by the following Ordinary Differential Equations (ODE):
	\begin{equation}
	\begin{split}
	\frac{d u_{jk}}{d t} &= -\frac{\partial \mathcal{L}_{aug}\left ( \mU(t),\mV(t),\mW(t), \va(t) \right )}{\partial u_{jk}(t)}, \\
	\frac{d v_{ij}}{d t} &= -\frac{\partial \mathcal{L}_{aug}\left ( \mU(t),\mV(t),\mW(t), \va(t)  \right )}{\partial v_{ij}(t)}
	\end{split}
	\end{equation}
	for $i \in [d_{out}]$, $j \in [m]$ and $k \in [d_{in}]$. In ideal condition, the system enters equilibrium when $\frac{d u_{jk}}{d t}=\frac{d v_{ij}}{d t}=0$. To circumvent tractability issues, we seek $\epsilon$-approximate equilibrium, i.e. $\left | \frac{d u_{jk}}{d t} \right |<\epsilon$ and $ \left | \frac{d v_{ij}}{d t} \right |<\epsilon$ for a small $\epsilon$.

	\subsection{Revisiting Reaction-Diffusion Model\cite{turing1952chemical}}
	\label{rdm}
	We focus on two body morphogenesis though it may be applied generally to many bodies upon further investigation. Here, two bodies refer to two layers of generator network. There are $2m$  differential equations governing the reaction ($\mathfrak{R}$) and diffusion ($\mathfrak{D}$) dynamics of such a complex system:
	\begin{equation}
	\begin{split}
	\frac{d \vu_j}{d t} &= \mathfrak{R}^{\vu}_{j}\left ( \vu_j, \vv_j \right ) + \mathfrak{D}^{\vu}_{j}\left ( \nabla^2 \vu_j \right ), \\
	\frac{d \vv_j}{d t} &= \mathfrak{R}^{\vv}_{j}\left ( \vu_j, \vv_j \right ) + \mathfrak{D}^{\vv}_{j}\left ( \nabla^2 \vv_j \right ),
	\end{split}
	\label{rd-eq}
	\end{equation}
	where $j=1,2,\dots,m$. Here, $m$ denotes the total number of neurons in the hidden layer. In the current setup, $\vu_j = \left (u_{jk} \right )_{k=1}^{d_{in}}, u_{jk} \in \mathbb{R}$ and $\vv_j = \left (v_{ij} \right )_{i=1}^{d_{out}}, v_{ij} \in \mathbb{R}$. Thus, $\frac{d \vu_j }{d t} = \left ( \frac{d u_{jk}}{d t} \right )_{k=1}^{d_{in}}$ and $\frac{d \vv_j }{d t} = \left ( \frac{d v_{ij}}{d t}\right )_{i=1}^{d_{out}}$. In the current analogy, each neuron represents a morphogen as it fulfills the fundamental requirements of Turing pattern formation. For better understanding, we have grouped those in hidden layer to one entity ($\vu_j$) and top layer to another entity ($\vv_j$). Among several major advantages of RD systems, a few that are essential to the present body of analysis are separability, stability and strikingly rich spatio-temporal dynamics. Later parts of this paper will focus on deriving suitable expressions for the reaction and diffusion term.

	\subsection{Pseudo-Reaction-Diffusion Model}
	\label{prdm}
	The analogy that has been made with RD systems in the foregoing analysis may be rather confusing to some readers. The succeeding analysis is intended to clarify some of these concerns. In the traditional setting, diffusion terms are limited to the Laplacian of the corresponding morphogens. In the present account however, the diffusibility of one morphogen depends on the other morphogens, and hence the term \textit{pseudo-reaction-diffusion}. Since later discoveries identified the root cause of pattern formation to be a short range positive feedback and a long range negative feedback~\cite{meinhardt1974applications,meinhardt2000pattern,rauch2004role}, a system with adversarial interaction is fairly a pseudo-reaction-diffusion model.

	\section{Theoretical Analysis}
	\label{ta}
	First, we study symmetry and homogeneity in a simplified setup. In this regard, the separability property allows us to choose a scalar network, i.e., $d_{out}=1$ and fix the second layer weights. There are $2m$ morphogens in the hidden layer itself making it a critically important analysis from mathematics perspective. Even with this simplification, the network is still non-convex and non-smooth\footnote{We choose to fix the weights of the second layer because the network becomes convex and smooth if we fix the weights of the first layer. It motivates us to make this choice which is not far from practice and allows us to simplify the expressions.}. The network architecture then becomes:
	\begin{equation}
	f\left ( \mU,\vv,\vx \right ) = \frac{1}{\sqrt{m}} \sum_{j=1}^{m} v_j \sigma\left ( u_j^T x \right ) = \frac{1}{\sqrt{m}} \vv^T \sigma\left ( \mU \vx \right ).
	\end{equation}
	Our goal is to minimize
	\begin{equation}
	\mathcal{L}_{sup}\left ( \mU, \vv \right ) = \sum_{p=1}^{n} \frac{1}{2} \left ( f\left ( \mU,\vv,\vx_p \right )- y_p\right )^2
	\label{sup_eq}
	\end{equation}
	in a supervised setting and $\mathcal{L}_{aug}\left ( \mU, \vv, \vw, \va \right ) $
	\begin{equation}
	\begin{split}
	= \sum_{p=1}^{n} \frac{1}{2} \left ( f\left ( \mU,\vv,\vx_p \right )- y_p\right )^2 -  \frac{1}{\sqrt{m}} \sum_{p=1}^{n} \va^T ~\sigma \left ( \vw \left ( f\left ( \mU,\vv,\vx_p \right ) \right )  \right )
	\end{split}
	\label{aug_eq}
	\end{equation}
	in an adversarial setting. The architecture of adversary is simplified to $g\left ( w,a,y \right ) = \frac{1}{\sqrt{m}} \sum_{j=1}^{m} a_j \sigma\left ( w_j y \right )$. In the adversarial setting, this problem can be related to min-max optimization in non-convex-non-concave setting. We follow the definition of Gram matrix from \citeauthor{du2018gradient}
	
	\textbf{Definition 1.} \textit{Define Gram matrix $\mathcal{H}^\infty \in \mathcal{R}^{n\times n}$. Each entry of $\mathcal{H}^\infty$ is computed by $\mathcal{H}_{ij}^\infty = \mathbb{E}_{u\sim \mathcal{N}\left ( 0,I \right )} \left [ x_i^T x_j 1_{\left \{ u^T x_i \geq 0, u^T x_j \geq 0 \right \}} \right ]$. }
	
	Let us recall the following assumption which is crucial for the analysis in this paper. 
	
	\textbf{Assumption 1.} \textit{We assume $\lambda_0 \triangleq \lambda_{min}\left ( \mathcal{H}^\infty \right ) > 0$ which means that $\mathcal{H}^\infty $ is a positive definite matrix.}
	
	The Gram matrix has several important properties~\cite{tsuchida2018invariance,xie2017diverse}. One interesting property that justifies \textbf{Assumption 1} is given by \citeauthor{du2018gradient}: \textit{If no two inputs are parallel, then the Gram matrix is positive definite.} This is a valid assumption as very often we do not rely on a training dataset that contains too many parallel samples.

	\subsection{Warm-Up: Reaction Without Diffusion}
	\label{wu}
	Before stating the main result, it is useful to get familiarized with the arguments of warm-up exercise. 
	
	\textbf{Theorem 1.} (Symmetry and Homogeneity) \textit{Suppose \textbf{Assumption 1} holds. Let us i.i.d. initialize $u_j \sim \mathcal{N}\left(0,I\right)$ and sample $v_j$ uniformly from $\left \{ +1,-1\right \}$ for all $j \in [m]$. If we choose $\left \|x_p\right \|_2=1$ for $p\in [n]$, then we obtain the following with probability at least $1-\delta$:
		\begin{equation*}
		\begin{split}
		\left \| \vu_j(t) - \vu_j(0) \right \|_2 & \leq \mathcal{O}\left ( \frac{n^{3/2}}{m^{1/2}\lambda_0 \delta} \right ),\\ \left \| \mU(t) - \mU(0) \right \|_F & \leq \mathcal{O}\left ( \frac{n^{3/2}}{\lambda_0 \delta} \right ).
		\end{split}
		\end{equation*}}
	
	\textit{Proof.} We begin proof sketch with the following lemma.  
	
	\textbf{Lemma 1.} \textit{If we i.i.d initialize $u_{jk}\sim \mathcal{N}\left (0,1 \right )$ for $j \in [m]$ and $k\in [d_{in}]$, then with probability at least $(1-\delta)$, $u_{jk}$ induces a symmetric and homogeneously distributed matrix $U$ at initialization within a ball of radius $\zeta \triangleq \frac{2\sqrt{m d_{in}}}{\sqrt{2\pi}\delta}$.}
	
	\textit{Proof.} Using the law of large numbers, it is trivial to prove symmetry and homogeneity since Gaussian distribution has a symmetric density function. We defer the proof of upper bound to Appendix.
	
	Next, we prove how supervised cost helps maintain symmetry and homogeneity. Since $\mU$ is initially symmetric and homogeneously distributed within $\zeta$ according to \textbf{Lemma~1}, the problem is now reduced to show that $\mU(t)$ lies in the close proximity of $\mU(0)$. We remark three crucial observations from \citeauthor{du2018gradient} that are essential to our analysis.

	\textbf{Remark 1.} \textit{Suppose $\left \| \vu_j - \vu_j(0) \right \|_2\leq \frac{c\delta \lambda_0}{n^2} \triangleq R$ for some small positive constant $c$. In the current setup, the Gram matrix $\mathcal{H} \in \mathbb{R}^{n\times n}$ defined by 
		\begin{equation*}
		\mathcal{H}_{ij} = \vx_i^T \vx_j \frac{1}{m}\sum_{r=1}^{m} 1_{\left \{\vu_r^T \vx_i \geq 0, \vu_r^T \vx_j \geq 0\right \}}
		\end{equation*}	
		satisfies $
		\left \| \mathcal{H} -\mathcal{H}(0) \right \|_2 \leq \frac{\lambda_0}{4}~\text{ and } \lambda_{min}\left ( \mathcal{H} \right ) \geq \frac{\lambda_0}{2}.$}
	
	\textbf{Remark 2.} \textit{With Gram matrix $\mathcal{H}(t)$, the prediction dynamics, $z(t) = f\left ( \mU(t), \vv(t), \vx \right )$ are governed by the following ODE:
		\begin{equation*}
		\frac{d \vz(t)}{d t} = \mathcal{H}(t)\left ( \vy-\vz(t) \right ).
		\end{equation*}}
	
	\textbf{Remark 3.} \textit{ For $\lambda_{min} \left ( \mathcal{H}(t) \right) \geq \frac{\lambda_0}{2}$, we have 
		\begin{equation*}
		\left \| \vz(t) - \vy \right \|_2 \leq \exp{\left (-\frac{\lambda_0}{2} t \right )} \left \| \vz(0) - \vy \right \|_2.
		\end{equation*}}
	
	Now, for $0\leq s \leq t$,
	\begin{equation}
	\begin{split}
	\left \| \frac{d \vu_j(s)}{d s} \right \|_2 & = \left \| \frac{\partial \mathcal{L}_{sup}\left ( \mU, \vv \right )}{\partial \vu_j(s)} \right \|_2
	= \left \| \sum_{p=1}^{n} \left ( z_p(s) - y_p \right ) \frac{\partial z_p(s)}{\partial \vu_j(s)} \right \|_2 \\
	& = \left \| \sum_{p=1}^{n} \left ( z_p(s) - y_p \right ) \frac{1}{\sqrt{m} } v_j 1_{\left \{ \vu_j(s)^T \vx_p \geq 0 \right \}} \vx_p \right \|_2.
	\end{split}
	\end{equation}
	By triangle inequality, 
	\begin{equation}
	\left \| \frac{d \vu_j(s)}{d s} \right \|_2 \leq \sum_{p=1}^{n} \left \| \left ( z_p(s) - y_p \right ) \frac{1}{\sqrt{m} } v_j 1_{\left \{ u_j(s)^Tx_p \geq 0 \right \}} x_p \right \|_2.
	\end{equation}
	Using the classical inequality of Cauchy-Schwarz, $\left \|x_p \right \|_2 = 1$, $|v_j|=1$ and \textbf{Remark 3}, we get
	\begin{equation}
	\begin{split}
	\left \| \frac{d \vu_j(s)}{d s} \right \|_2 & \leq \sum_{p=1}^{n} \frac{1}{\sqrt{m} } \left |\left ( z_p(s) - y_p \right )  \right |  \left |v_j  \right |  \left \|  x_p \right \|_2 \\&=  \frac{1}{\sqrt{m} } \sum_{p=1}^{n} \left |\left ( z_p(s) - y_p \right )  \right | \\
	& \leq \frac{\sqrt{n}}{\sqrt{m}}\left \| \vz(s) - \vy \right \|_2 \\&\leq  	\frac{\sqrt{n}}{\sqrt{m}} \exp{\left (-\frac{\lambda_0}{2} s \right )} \left \| \vz(0) - \vy \right \|_2.
	\end{split}
	\label{warm_up_eq}
	\end{equation}
	By integral form of Jensen's inequality, the distance from initialization can be bounded by
	\begin{equation}
	\begin{split}
	\left \| \vu_j(t) - \vu_j(0) \right \|_2 &= \left \| \int_{0}^{t} \frac{d \vu_j(s)}{d s}  d s\right \|_2 \leq  \int_{0}^{t} \left \|\frac{d \vu_j(s)}{d s} \right \|_2  d s \\ & \leq  \frac{\sqrt{n}}{\sqrt{m}} \int_{0}^{t} \exp{\left (-\frac{\lambda_0}{2} s \right )} \left \| \vz(0) - \vy  \right \|_2 d s \\&\leq \frac{2\sqrt{n} \left \| \vz(0) - \vy  \right \|_2}{\sqrt{m} \lambda_0} \left ( 1- \exp{ \left (-\frac{\lambda_0}{2} t  \right )} \right ).
	\end{split}
	\end{equation}
	Since $\exp{\left (-\frac{\lambda_0}{2} t \right )}$ is a decreasing function of $t$, the above expression simplifies to 
	\begin{equation}
	\left \| \vu_j(t) - \vu_j(0) \right \|_2 \leq \frac{2\sqrt{n} \left \| \vz(0) - \vy  \right \|_2}{\sqrt{m} \lambda_0}.
	\end{equation}
	Using Markov's inequality, with probability at least $1-\delta$, we get
	\begin{equation}
	\begin{split}
	\left \| \vu_j(t) - \vu_j(0) \right \|_2 & \leq \frac{2\sqrt{n} \mathbb{E}\left [\left \| \vz(0) - \vy  \right \|_2  \right ]}{\sqrt{m} \lambda_0 \delta} \\&\leq  \mathcal{O}\left ( \frac{n^{3/2}}{m^{1/2} \lambda_0 \delta} \right ).
	\end{split}
	\end{equation}
	Now, we can bound the distance from initialization.
	\begin{equation}
	\begin{split}
	\left \| \mU(t) - \mU(0) \right \|_F & = \left ( \sum_{j=1}^{m} \sum_{k=1}^{d_{in}} \left | u_{jk}(t) - u_{jk}(0) \right |^2  \right )^{1/2} \\&\leq \left ( \sum_{j=1}^{m} \left \| \vu_j(t) - \vu_j(0) \right \|_2^2 \right )^{1/2} \\ & \leq \left ( \sum_{j=1}^{m} \frac{4n \left (\mathbb{E}\left[ \left \| \vz(0) - \vy \right \|_2\right]\right )^2}{m \lambda_0^2 \delta^2} \right )^{1/2} \\&\leq  \frac{2\sqrt{n} \mathbb{E}\left[\left \| \vz(0) - \vy \right \|_2 \right]}{ \lambda_0 \delta} \leq \mathcal{O}\left ( \frac{n^{3/2}}{\lambda_0 \delta } \right ),
	\end{split}
	\end{equation}
	which finishes the proof.$ \hfill \square$

	\subsection{Main Result: Reaction With Diffusion}
	\label{mr}
	
	To limit the capacity of a discriminator, it is often suggested to enforce a Lipschitz constraint on its parameters. While gradient clipping has been quite effective in this regard~\cite{arjovsky2017wasserstein}, recent success in adversarial training owes in part to gradient penalty~\cite{gulrajani2017improved}. We remark that min-max optimization under non-convexity and non-concavity is considered NP-hard to find a stationary point~\cite{lei2019sgd}. Therefore, it is necessary to make certain assumptions about discriminator, such as Lipschitz constraint, regularization and structure of the network. Different from one layer generator and quadratic discriminator~\cite{lei2019sgd}, we study two layer networks with ReLU activations and rely on gradient penalty to limit its expressive power. In the simplified theoretical analysis, we assume $\left \| \vw \right \|_2 \leq L$ for a small constant $L > 0 $.

	\textbf{Theorem 2.} (Breakdown of Symmetry and Homogeneity) \textit{Suppose \textbf{Assumption 1} holds. Let us i.i.d. initialize $u_j, w_r \sim \mathcal{N}\left(0,I\right)$ and sample $v_j, a_r$ uniformly from $\left \{ +1,-1\right \}$ for $j, r \in [m]$. Let $\left \|x_p\right \|_2=1$ for all $p\in [n]$. If we choose $\left \| \vw \right \|_2 \leq L \leq \mathcal{O}\left ( \frac{\epsilon\sqrt{m}}{\kappa n \sqrt{2 \log\left ( 2/\delta \right )}} \right )$, $\kappa = \mathcal{O}(\kappa^{\infty})$ where $\kappa^{\infty}$denotes the condition number of $\mathcal{H}^\infty$, and  define $\mu \triangleq \frac{Ln\sqrt{2  \log\left ( 2/\delta \right )}}{\sqrt{m}}$, then with probability at least $1-\delta$, we obtain the following\footnote{Refer to Appendix for experimental evidence and further discussion on breakdown of symmetry and homogeneity.}:
		\begin{equation*}
		\begin{split}
		\left \| \vu_j(t) - \vu_j(0) \right \|_2 &\leq \mathcal{O}\left ( \frac{n^{3/2}}{\sqrt{m}\lambda_0 \delta} + \left ( \frac{\mu\left( 1+ \kappa \sqrt{n}\right)}{\sqrt{m}} \right ) t \right ), \\ \left \| \mU(t) - \mU(0) \right \|_F &\leq  \mathcal{O}\left ( \frac{n^{3/2}}{\lambda_0 \delta} +  \mu\left( 1+ \kappa \sqrt{n}\right) t\right ).
		\end{split}
		\end{equation*}}
	\textit{Proof.} We sketch the proof of the main result as following.	
	
	\subsubsection{Reaction Term}
	\label{rt}
	For $0\leq s \leq t$ in augmented objective as given by equation~(\ref{aug_eq}), we get
	\begin{equation}
	\begin{split}
	\left \| \frac{d \vu_j(s)}{d s} \right \|_2 & = \left \| \frac{\partial \mathcal{L}_{aug}\left ( \mU, \vv, \vw, \va \right )}{\partial \vu_j(s)} \right \|_2 \\&= \left \| \frac{\partial \mathcal{L}_{sup}\left ( \mU, \vv \right )}{\partial \vu_j(s)} - \frac{\partial }{\partial \vu_j(s)}\sum_{p=1}^{n} g\left( \vw,a, z_p\right) \right \|_2 \\
	& \leq \underset{\text{Triangle inequality}}{\underbrace{ \left \| \frac{\partial \mathcal{L}_{sup}\left ( \mU, \vv \right )}{\partial \vu_j(s)} \right \|_2 + \left \| \frac{\partial }{\partial \vu_j(s)}\sum_{p=1}^{n} g\left( \vw,a, z_p\right)  \right \|_2}}.
	\end{split}
	\end{equation}
	We start our analysis by first deriving an asymptotic upper bound of the supervised part. Then, we shift our focus to the augmented part which essentially constitutes the adversary. 
	
	\textbf{Lemma 2.} \textit{In contrast to \textbf{Remark 2}, the prediction dynamics in adversarial regularization are governed by the following ODE: 
		\begin{equation}
		\frac{d \vz(t)}{d t} = \mathcal{H}(t)\left ( \vy-\vz(t) \right ) + \mathcal{H}(t) \nabla_{\vz(t)}g(\vw(t),\va(t),\vz(t)).
		\end{equation}}
	
	\textit{Proof.} The above ODE is obtained by analyzing the dynamics as following:
	\begin{equation}
	\begin{split}
	&\frac{d z_p(t)}{d t} = \sum_{j=1}^{m} \left \langle \frac{\partial f\left ( \mU, \vv, \vx_p \right )}{\partial \vu_j(t)}, \frac{d \vu_j(t)}{d t} \right \rangle\\
	& = \underset{\mathcal{A}}{\underbrace{\sum_{j=1}^{m} \left \langle \frac{\partial f\left ( \mU, \vv, \vx_p \right )}{\partial \vu_j(t)}, \frac{1}{\sqrt{m}}\sum_{q=1}^{n}\left ( y_q - z_q \right ) v_j \vx_q 1_{\left \{ \vu_j^T \vx_q \geq 0 \right \}}\right \rangle}} \\ & \hspace{0.5cm} + \underset{\mathcal{B}}{\underbrace{\sum_{j=1}^{m} \left \langle \frac{\partial f\left ( \mU, \vv, \vx_p \right )}{\partial \vu_j(t)}, \frac{1}{m}\sum_{q=1}^{n} \sum_{r=1}^{m} a_r w_r v_j \vx_q 1_{\left \{ w_r z_q \geq 0, \vu_j^T \vx_q \geq 0 \right \}}  \right \rangle}}.
	\end{split}
	\end{equation}
	Following arguments of the warm-up exercise, the first part can be simplified as:
	\begin{equation}
	\begin{split}
	\mathcal{A} & \coloneqq \sum_{j=1}^{m} \left \langle \frac{1}{\sqrt{m}} v_j \vx_p 1_{\left \{ \vu_j^T \vx_p \geq 0 \right \}}, \frac{1}{\sqrt{m}}\sum_{q=1}^{n}\left ( y_q - z_q \right ) v_j \vx_q 1_{\left \{ \vu_j^T \vx_q \geq 0 \right \}}\right \rangle \\
	& = \sum_{q=1}^{n} \left ( y_q - z_q \right ) \vx_p^T \vx_q \frac{1}{m} \sum_{j=1}^{m} 1_{\left \{ \vu_j^T \vx_p \geq 0, \vu_j^T \vx_q \geq 0  \right \}}\\
	& \triangleq \sum_{q=1}^{n} \left ( y_q - z_q (t) \right ) \mathcal{H}_{pq}(t),
	\end{split}
	\end{equation}
	where $\mathcal{H}_{pq}(t)$ denotes the elements of Gram matrix $\mathcal{H}(t)$ defined by
	\begin{equation}
	\mathcal{H}_{pq}(t) = \vx_p^T \vx_q \frac{1}{m} \sum_{j=1}^{m} 1_{\left \{ \vu_j^T \vx_p \geq 0, \vu_j^T \vx_q \geq 0  \right \}}.
	\end{equation}
	Using the predefined Gram matrix, the second part can be simplified as:
	\begin{equation}
	\begin{split}
	\mathcal{B} & \coloneqq \sum_{j=1}^{m} \left \langle \frac{1}{\sqrt{m}} v_j \vx_p 1_{\left \{ \vu_j^T \vx_p \geq 0 \right \}}, \frac{1}{m}\sum_{q=1}^{n} \sum_{r=1}^{m} a_r w_r v_j \vx_q 1_{\left \{ w_r z_q \geq 0, \vu_j^T \vx_q \geq 0 \right \}}  \right \rangle \\
	& = \sum_{q=1}^{n}  \underset{\nabla_\vz g}{\underbrace{\left (\frac{1}{\sqrt{m}} \sum_{r=1}^{m} a_r w_r 1_{\left \{ w_r z_q \geq 0 \right \}}  \right )}} \vx_p^T \vx_q \frac{1}{m} \sum_{j=1}^{m} 1_{\left \{ \vu_j^T \vx_p \geq 0, \vu_j^T \vx_q \geq 0  \right \}} \\
	& \triangleq \sum_{q=1}^{n} \frac{\partial g \left ( \vw, \va, z_q \right )}{\partial z_q} \mathcal{H}_{pq}(t)
	\end{split}
	\end{equation}
	Thus, the prediction dynamics are governed by
	\begin{equation}
	\begin{split}
	\frac{d z_p(t)}{d t} &= \sum_{q=1}^{n} \left ( y_q - z_q (t) \right ) \mathcal{H}_{pq}(t) \\&+  \sum_{q=1}^{n} \frac{\partial g \left ( \vw(t), \va(t), z_q(t) \right )}{\partial z_q(t)} \mathcal{H}_{pq}(t).
	\end{split}
	\end{equation}
	Rearranging the above expression in matrix form, we get the statement of \textbf{Lemma 2}. \hfill $\square$ \\
	
	\textbf{Lemma 3.} (Hoeffding's inequality, two sided~\cite{vershynin2018high}) \textit{Suppose $\va = (a_1, a_2,\dots, a_m) \in \left \{\pm 1 \right \}^{m}$ be a collection of independent symmetric Bernoulli random variables, and $\vw=(w_1,w_2,\dots,w_m) \in \mathbb{R}^{m}$. Then, for any $t>0$, we have 
		\begin{equation}
		\mathbb{P}\left \{ \left | \sum_{r=1}^{m} a_r w_r \right | \geq t \right \}\leq 2 \exp \left ( -\frac{t^2}{2\left \| \vw \right \|_2^2} \right ).
		\end{equation}}
	With probability at least $1-\delta$, we get the following bound using two-sided Hoeffding's inequality:
	\begin{equation}
	\left | \sum_{r=1}^{m} a_r w_r \right | \leq \left \| \vw \right \|_2 \sqrt{2  \log\left ( \frac{2}{\delta} \right )}.
	\label{hding}
	\end{equation}
	Now, the distance from true labels can be bounded by
	\begin{equation}
	\begin{split}
	&\frac{d}{d t} \left \| \vz(t) - \vy \right \|_2^2 \\& = 2 \left \langle  \vz(t) - \vy, \frac{d \vz(t)}{d t} \right \rangle\\
	& = 2 \left \langle  \vz(t) - \vy,  - \mathcal{H}(t)\left ( \vz(t) - \vy \right )  \right \rangle \\& \hspace{0.5cm}+ 2 \left \langle  \vz(t) - \vy, \mathcal{H}(t) \nabla_{\vz(t)}g(\vw(t),\va(t),\vz(t)) \right \rangle\\
	\end{split}
	\end{equation}
	
	\textbf{Lemma 4.} \textit{Suppose \textbf{Assumption 1} holds. If we denote $\lambda_{\max}\left ( \mathcal{H}^{\infty} \right )$ by $\lambda_1^{\infty}$, then $\lambda_{\max}\left ( \mathcal{H} \right ) \leq \frac{\lambda_1}{2} \triangleq \lambda_1^{\infty} + \frac{\lambda_0}{2}$.
	}
	\textit{Proof.} As the proof is relatively simpler, we defer it to appendix.
	
	Since $\lambda_{\min}(\mathcal{H}) \geq \frac{\lambda_0}{2}$ (\textbf{Remark 1}) and $\lambda_{\max}(\mathcal{H}) \leq \frac{\lambda_1}{2}$ (\textbf{Lemma 4}), we get
	\begin{equation}
	\begin{split}
	&\frac{d}{d t} \left \| \vz(t) - \vy \right \|_2^2 \\& \leq  -\lambda_0 \left \| \vz(t) - \vy \right \|_2^2 \\& \hspace{0.5cm}+ \lambda_1 \left \langle  \vz(t) - \vy, \nabla_{\vz(t)}g(\vw(t),\va(t),\vz(t)) \right \rangle\\
	& \leq -\lambda_0 \left \| \vz(t) - \vy \right \|_2^2 \\& \hspace{0.5cm}+ \lambda_1 \underset{\text{Cauchy-Schwarz inequality}}{\underbrace{ \left \|  \vz(t) - \vy\right \|_2 \left \| \nabla_{\vz(t)}g(\vw(t),\va(t),\vz(t)) \right \|_2}}\\
	& \leq  -\lambda_0 \left \| \vz(t) - \vy \right \|_2^2 \\& \hspace{0.5cm}+ \lambda_1  \left \| \vz(t) - \vy\right \|_2 \left \| \nabla_{\vz(t)}g(\vw(t),\va(t),\vz(t))  \right \|_1\\
	& \leq  -\lambda_0 \left \| \vz(t) - \vy \right \|_2^2 \\& \hspace{0.5cm} + \lambda_1 \left \|  \vz(t) - \vy\right \|_2 \sum_{q=1}^{n}  \left | \frac{1}{\sqrt{m}}\sum_{r=1}^{m} a_r w_r 1_{\left \{ w_r z_q \geq 0 \right \}}   \right |\\
	& \leq  -\lambda_0 \left \| \vz(t) - \vy \right \|_2^2 \\& \hspace{0.5cm} + \lambda_1 \left \|  \vz(t) - \vy\right \|_2  \frac{n}{\sqrt{m}}  \left | \sum_{r=1}^{m} a_r w_r \right |
	\end{split}
	\label{cau_sch}
	\end{equation}
	Substituting equation~(\ref{hding}) in equation~(\ref{cau_sch}), we get
	\begin{equation}
	\begin{split}
	& \frac{d}{d t} \left \| \vz(t) - \vy \right \|_2^2 \\& \leq -\lambda_0 \left \| \vz(t) - \vy \right \|_2^2 + \lambda_1 \left \|  \vz(t) - \vy\right \|_2  \frac{n}{\sqrt{m}} \left \| \vw \right \|_2 \sqrt{2  \log\left ( \frac{2}{\delta} \right )}\\
	& \leq -\lambda_0 \left \| \vz(t) - \vy \right \|_2^2 +  \frac{\lambda_1 Ln \sqrt{2  \log\left ( \frac{2}{\delta} \right )}}{\sqrt{m}} \left \|  \vz(t) - \vy\right \|_2.
	\end{split}
	\end{equation}
	Let us define $\mu \triangleq \frac{Ln\sqrt{2  \log\left ( \frac{2}{\delta} \right )}}{\sqrt{m}}$.  Then,
	\begin{equation}
	\begin{split}
	\frac{d}{d t} \left \| \vz(t) - \vy \right \|_2^2 \leq -\lambda_0 \left \| \vz(t) - \vy \right \|_2^2 +  \lambda_1 \mu \left \|  \vz(t) - \vy\right \|_2 
	\end{split}
	\end{equation}
	The above non-linear ODE is a special Bernoulli Differential Equation~(BDE)\footnote{A Bernoulli differential equation is an ODE of the form $\frac{d x(t)}{d t} + P(t) x(t) = Q(t)x^n(t)$ for $n\in \mathbb{R}\backslash\left \{0,1 \right \}.$ } which has known exact solutions~\cite{bernoulli1695explicationes}. For simplicity, let us suppose $\psi = \left \| \vz(t) - \vy \right \|_2^2$. Now, 
	\begin{equation}
	\begin{split}
	\frac{d  \psi}{d t} \leq -\lambda_0 \psi +  \lambda_1 \mu \psi^{1/2} 
	\end{split}
	\end{equation}
	Substituting $\psi= \varphi^2$, the BDE is reduced to an Initial Value Problem (IVP): $\frac{d  \varphi}{d t} + \frac{\lambda_0}{2} \varphi \leq  \frac{\lambda_1}{2} \mu $. By substituting $\varphi = \nu \zeta$, the IVP is decomposed into two linear ODEs of the form $\frac{d \nu}{d t} + \frac{\lambda_0}{2} \nu = 0$ and $\nu \frac{d \zeta}{d t} - \frac{\lambda_1}{2} \mu  = 0$. Since these ODEs have separable forms, for arbitrary constants $C_1$ and $C_2$, we get 
	\begin{equation}
	\begin{split}
	\nu = C_1 \exp{\left(-\frac{\lambda_0 t}{2} \right)},~\zeta = C_2 + \frac{\kappa \mu}{C_1} \exp{\left( \frac{\lambda_0 t}{2}  \right)},
	\end{split}
	\end{equation}
	where $\kappa=\frac{\lambda_1}{\lambda_0} = \frac{2\left ( \lambda_1^{\infty} + \frac{\lambda_0}{2} \right )}{\lambda_0} = \mathcal{O}(\kappa^{\infty})$.  Here, $\kappa^{\infty} $ is the condition number of $\mathcal{H}^{\infty}$. Thus, the solution of the BDE is given by $\psi = \varphi^2 = \left( C \exp{\left(-\frac{\lambda_0t}{2}\right)} + \kappa \mu\right)^2$ for another constant $C$. Using initial value of $\psi$, we get the exact solution:
	\begin{equation}
	\begin{split}
	\left \| \vz(t) - \vy \right \|_2 \leq \left(\left \| \vz(0) - \vy \right \|_2 - \kappa \mu  \right) \exp{\left(-\frac{\lambda_0}{2}t\right)} + \kappa \mu.
	\end{split}
	\label{pred_dynamics}
	\end{equation}
	From equation~(\ref{warm_up_eq}) in the warm-up exercise, we know for $0\leq s \leq t$,
	\begin{equation}
	\begin{split}
	\left \| \frac{\partial \mathcal{L}_{sup}\left ( \mU, \vv \right )}{\partial \vu_j(s)} \right \|_2 &\leq \frac{\sqrt{n}}{\sqrt{m}}\left \| \vz(s) - \vy \right \|_2.
	\end{split}
	\end{equation}
	Now, substituting equation~(\ref{pred_dynamics}), we get
	\begin{equation}
	\begin{split}
	&\left \| \frac{\partial \mathcal{L}_{sup}\left ( \mU, \vv \right )}{\partial \vu_j(s)} \right \|_2 \\ &\leq \frac{\sqrt{n}}{\sqrt{m}} \left(\left \| \vz(0) - \vy \right \|_2 - \kappa \mu  \right) \exp{\left(-\frac{\lambda_0}{2}s\right)} + \frac{\sqrt{n}}{\sqrt{m}} \kappa \mu.
	\end{split}
	\end{equation}
	Therefore, the reaction dynamics are given by
	\begin{equation}
	\begin{split}
	\mathfrak{R}_j^u(\vu_j(t)) \leq \frac{\sqrt{n}}{\sqrt{m}} \left(\left \| \vz(0) - \vy \right \|_2 - \kappa \mu  \right) \exp{\left(-\frac{\lambda_0}{2}t\right)} + \frac{\sqrt{n}}{\sqrt{m}} \kappa \mu.
	\end{split}
	\end{equation}

	\subsubsection{Diffusion Term}
	\label{dt}
	The augmented part on the other hand becomes:
	\begin{equation}
	\begin{split}
	&\left \| \frac{\partial }{\partial \vu_j(s)}\sum_{p=1}^{n} g\left( \vw,a, z_p\right)  \right \|_2 \\ & =  \left \| \sum_{p=1}^{n}  \sum_{r=1}^{m} \frac{1}{\sqrt{m}}a_r 1_{\left \{ w_r z_p \geq 0 \right \}} w_r \frac{1}{\sqrt{m}} v_j 1_{\left \{ \vv_j^T \vx_p \geq 0 \right \}} \vx_p  \right \|_2.
	\end{split}
	\end{equation}
	By Triangle and Cauchy-Schwarz inequality, we get
	\begin{equation}
	\begin{split}
	&\left \| \frac{\partial }{\partial \vu_j(s)}\sum_{p=1}^{n} g\left( \vw,a, z_p\right)  \right \|_2 \\& \leq  \frac{1}{m} \sum_{p=1}^{n}  \left \| v_j 1_{\left \{ \vv_j^T \vx_p \geq 0 \right \}}  \vx_p \sum_{r=1}^{m} a_r w_r  1_{\left \{w_r z_p \geq 0\right \}}  \right \|_2\\
	&  \leq  \frac{1}{m} \sum_{p=1}^{n}   |v_j|   \left \|\vx_p\right \|_2 \left |\sum_{r=1}^{m} a_r w_r  \right |   \\
	& \leq  \frac{1}{m} \sum_{p=1}^{n}  \left |\sum_{r=1}^{m} a_r w_r  \right |  	
	\end{split}
	\label{tring_cauchy1}
	\end{equation}
	Substituting equation~(\ref{hding}) in equation~(\ref{tring_cauchy1}), we arrive at the following inequality:
	\begin{equation}
	\begin{split}
	\left \| \frac{\partial }{\partial \vu_j(s)}\sum_{p=1}^{n} g\left( \vw,a, z_p\right)  \right \|_2 	& \leq  \frac{1}{m} \sum_{p=1}^{n}  \left \| \vw \right \|_2 \sqrt{2  \log\left ( \frac{2}{\delta} \right )}\\ & \leq \frac{Ln\sqrt{2  \log\left ( \frac{2}{\delta} \right )}}{m} =  \mathcal{O}\left( \frac{\mu}{\sqrt{m}}	\right).
	\end{split}
	\end{equation}
	Thus, the diffusion dynamics are given by
	\begin{equation}
	\begin{split}
	\mathfrak{D}_j^u (\vu_j(t)) \leq\frac{Ln\sqrt{2  \log\left ( \frac{2}{\delta} \right )}}{m} .
	\end{split}
	\label{dju}
	\end{equation}
	Now integrating the gradients over $0\leq s\leq t$, 
	\begin{equation}
	\begin{split}
	&\left \| \vu_j(t) - \vu_j(0) \right \|_2 \leq \int_{0}^{t} \left \| \frac{d \vu_j(s)}{d s} \right\|_2 d s\\& \leq  \int_{0}^{t} \frac{\sqrt{n}}{\sqrt{m}}  \left(\left \| \vz(0) - \vy \right \|_2 - \kappa \mu  \right) \exp\left ( -\frac{\lambda_0}{2}  s \right ) d s \\&\hspace{3cm}+ \int_{0}^{t} \frac{\mu \left( 1+ \kappa \sqrt{n}\right)}{\sqrt{m}}  d s \\ 
	& \leq \frac{2 \sqrt{n}  \left(\left \| \vz(0) - \vy \right \|_2 - \kappa \mu  \right)}{\sqrt{m} \lambda_0 } \left ( 1-  \exp\left ( -\frac{\lambda_0}{2} t \right ) \right ) \\& \hspace{3cm}+  \left ( \frac{\mu\left( 1+ \kappa \sqrt{n}\right)}{\sqrt{m}} \right ) t. 
	\end{split}
	\end{equation}
	Using Markov's inequality, $\left \| \vz(0) - \vy \right \|_2 \leq \frac{\mathbb{E}\left [ \left \| \vz(0) - \vy \right \|_2\right ]}{\delta} = \mathcal{O}\left( \frac{n}{\delta}\right)$ with probability at least $1-\delta$. Thus, 
	\begin{equation}
	\begin{split}
	&\left \| \vu_j(t) - \vu_j(0) \right \|_2 \\& \leq \mathcal{O}\left ( \frac{n^{3/2}}{m^{1/2}\lambda_0 \delta} + \left ( \frac{\mu\left( 1+ \kappa \sqrt{n}\right)}{m^{1/2}} \right ) t \right ).
	\end{split}
	\label{u_exp_lin}
	\end{equation}
	Furthermore, the spatial grid of neurons satisfies:
	\begin{equation}
	\begin{split}
	\left \| \mU(t) - \mU(0) \right \|_F &\leq \sqrt{m} \left \| \vu_j(t) - \vu_j(0) \right \|_2 \\
	&\leq  \mathcal{O}\left ( \frac{n^{3/2}}{\lambda_0 \delta} +  \mu\left( 1+ \kappa \sqrt{n}\right) t\right ).
	\end{split}
	\end{equation}
	To circumvent tractability issues, it is common to seek an $\epsilon$-stationary point. As given by equation~(\ref{pred_dynamics}),  $\vz(t)$ in adversarial learning converges uniformly to an $\epsilon$-neighborhood of $\vy$ for any $t \geq T_0 \triangleq \frac{2}{\lambda_0} \log\left ( \frac{\left \| \vz(0) -\vy \right \|_2 - \kappa \mu}{\epsilon - \kappa \mu} \right )$. For finite time convergence, we need $\kappa \mu < \epsilon < \left \| \vz(0) -\vy \right \|_2$. The second inequality holds because we usually look for a solution where the error is better than what obtained during initialization. The first inequality gives the upper bound on gradient penalty, i.e., $L \leq \mathcal{O}\left ( \frac{\epsilon\sqrt{m}}{\kappa n \sqrt{2 \log\left ( 2/\delta \right )}} \right )$ by substituting the value of $\mu$. It is an important result in a sense that over-parameterized networks can still enjoy linear rate of convergence even under adversarial interaction. \hfill $\square$

	In a general configuration, \textbf{Remark 1} asserts that the induced Gram matrix is stable and satisfies our assumptions on eigen values as long as $\left \| \vu_j - \vu_j(0) \right \|\leq R$. Intuitively, this is satisfied when the points visited by gradient descent in adversarial learning lie within this $R$-ball. Formally, we need the following condition to be satisfied for finding the least expensive $\epsilon$-stationary point:
	\begin{equation}
	\begin{split}
	\mathcal{O}\left ( \frac{n^{3/2}}{m^{1/2}\lambda_0 \delta} + \left ( \frac{\mu\left( 1+ \kappa \sqrt{n}\right)}{m^{1/2}} \right ) T_0 \right ) \leq R.
	\end{split}
	\end{equation}
	Substituting $R=\frac{c\delta \lambda_0}{n^2}$ in the above expression, we get
	\begin{equation}
	\begin{split}
	m = \Omega\left ( \left ( \frac{n^{7/2}}{\lambda_0^2 \delta^2} + \frac{n^2 \mu \left( 1+ \kappa \sqrt{n}\right) T_0}{\lambda_0 \delta } \right )^2 \right ).
	\end{split}
	\end{equation}
	It is worth mentioning that the polynomial node complexity, $m = poly\left( n, \frac{1}{\lambda_0}, \frac{1}{\delta} \right)$ is also essential for finding an $\epsilon$-stationary point in sole supervision. By ignoring the diffusible factors, i.e., setting $\mu=0$, we recover the lower bound, $m = \Omega\left ( \frac{n^{7}}{\lambda_0^4 \delta^4} \right )$ in supervised learning.
	
	\section{Discussion of Insights from Analysis}
	\label{dia}
	
	A profound implication of this finding is that adversarial learning allows gradient descent to explore a large subspace in contrast to supervised learning where a tiny subspace around initialization is merely explored~\cite{gur2018gradient}. As a result, it offers the provision to exploit full capacity of network architectures by encouraging local interaction. In other words, the neurons in supervised learning do not interact with each other as much as they do in adversarial learning. By introducing the diffusible factors, it helps break the spatial symmetry and homogeneity in this tiny subspace. Due to more local interaction and diffusion, it exhibits patterns as a reminiscent of those observed in nature.  More importantly, this is consistent with the well-studied theory of pattern formation~\cite{turing1952chemical,meinhardt1982models,gray1984autocatalytic,rauch2004role}.

	The system of neurons is initially in a stable homogeneous condition due to non-diffusive elements in sole supervision. It is perturbed by irregularities introduced under the influence of an adversary. For a RD system, it is necessary that these irregularities are small enough, which otherwise would destabilize the whole system, and it may never converge to a reasonable solution. This is easily satisfied in over-parameterized networks as given by equation~(\ref{dju}). Thus, it is not unreasonable to suppose that adversarial interaction in augmented objective is the only one in which conditions are such to break the spatial symmetry. Different from strict RD systems, the diffusibility here does not directly depend on Laplacian of each morphogen. This is not uncommon because bell-like pattern formation in the skin of a zebrafish is a typical example where patterns emerge even when the system is different from the original Turing model~\cite{nakamasu2009interactions}. More importantly, it fits the description of short and long range feedback which indicates a similar mechanism must be involved in adversarial learning. This analogy provides positive support to the developed PRD theory.

	It is well known that randomly initialized gradient descent with over-parameterization finds solutions close to its initialization~\cite{du2018gradient,li2018learning,neyshabur2018role,nagarajan2019generalization}. The distance from initialization has helped unveil several mysteries of deep learning in part including the generalization puzzle and $\epsilon$-stationarity. We ask whether such implicit restriction to a tiny search space is a \textit{necessary condition} to achieve similar performance. The expressive power of a large network is not fully exploited by limiting the search space. This argument is supported by~\citeauthor{gulrajani2017improved} who show that the generator in WGAN with weight clipping~\cite{arjovsky2017wasserstein} fails to capture higher order moments. One reason for such behavior is the implicit restriction of discriminator weights to a tiny subspace around extremas due to weight clipping. It is resolved however by incorporating gradient penalty which allows exploration in a larger search space within clipping boundaries. In this regard, we provide both theoretical and empirical evidence that imposing such restriction is not a necessary condition. With over-parameterization, randomly initialized gradient descent can still find a global minimizer relatively farther from its initialization. It is possible because of adversarial interaction that helps introduce diffusible factors into the system. 
	
	\section{Conclusion \& Future Work}
	\label{conc}
	In this paper, we studied the evolutionary patterns formed in a system of neurons with adversarial interaction. We provided a theoretical justification and empirical evidence of Turing instability in a pseudo-reaction-diffusion model that underlie these phenomena. Furthermore, it was shown that randomly initialized gradient descent with over-parameterization could still enjoy exponentially fast convergence to an $\epsilon$-stationary point even under adversarial interaction. However, unlike sole supervision, it was found that the obtained solutions were not limited to a tiny subspace around initialization. It was observed that adversarial interaction helped in the breakdown of spatial symmetry and homogeneity which allowed exploration in a larger subspace. 
	
	While this work takes a step towards explaining non-homogeneous pattern formation due to adversarial interaction, it is far from being conclusive. Though diffusibility ensures more local interaction, it will certainly be interesting to synchronize neurons based on this observation in future.
	
	\section{Broader Impact}
	This paper investigates the underlying phenomena that may cause evolutionary patterns to emerge with the advent of adversarial interaction. By theoretical and empirical evidence, it tries to corroborate the developed pseudo-reaction-diffusion system. We believe this work does not present any forseeable societal consequence. 
	
	\bibliography{aaai21}

\begin{thebibliography}{44}
\providecommand{\natexlab}[1]{#1}
\providecommand{\url}[1]{\texttt{#1}}
\providecommand{\urlprefix}{URL }
\expandafter\ifx\csname urlstyle\endcsname\relax
  \providecommand{\doi}[1]{doi:\discretionary{}{}{}#1}\else
  \providecommand{\doi}{doi:\discretionary{}{}{}\begingroup
  \urlstyle{rm}\Url}\fi

\bibitem[{Allen-Zhu and Li(2020)}]{allen2020backward}
Allen-Zhu, Z.; and Li, Y. 2020.
\newblock Backward feature correction: How deep learning performs deep
  learning.
\newblock \emph{arXiv preprint arXiv:2001.04413} .

\bibitem[{Arjovsky, Chintala, and Bottou(2017)}]{arjovsky2017wasserstein}
Arjovsky, M.; Chintala, S.; and Bottou, L. 2017.
\newblock Wasserstein Generative Adversarial Networks.
\newblock In \emph{Proceedings of the 34th International Conference on Machine
  Learning}.

\bibitem[{Bernoulli(1695)}]{bernoulli1695explicationes}
Bernoulli, J. 1695.
\newblock Explicationes, Annotationes \& Additiones ad ea, quae in Actis sup.
  de Curva Elastica, Isochrona Paracentrica, \& Velaria, hinc inde memorata, \&
  paratim controversa legundur; ubi de Linea mediarum directionum, alliisque
  novis.
\newblock \emph{Acta Eruditorum} .

\bibitem[{Budzynski et~al.(2009)Budzynski, Budzynski, Evans, and
  Abarbanel}]{budzynski2009introduction}
Budzynski, T.~H.; Budzynski, H.~K.; Evans, J.~R.; and Abarbanel, A. 2009.
\newblock \emph{Introduction to quantitative EEG and neurofeedback: Advanced
  theory and applications}.
\newblock Academic Press.

\bibitem[{Du et~al.(2018)Du, Zhai, Poczos, and Singh}]{du2018gradient}
Du, S.~S.; Zhai, X.; Poczos, B.; and Singh, A. 2018.
\newblock Gradient Descent Provably Optimizes Over-parameterized Neural
  Networks.
\newblock In \emph{International Conference on Learning Representations}.

\bibitem[{Engin, Gen{\c{c}}, and Kemal~Ekenel(2018)}]{engin2018cycle}
Engin, D.; Gen{\c{c}}, A.; and Kemal~Ekenel, H. 2018.
\newblock Cycle-dehaze: Enhanced cyclegan for single image dehazing.
\newblock In \emph{Proceedings of the IEEE Conference on Computer Vision and
  Pattern Recognition Workshops}, 825--833.

\bibitem[{Frey and Dueck(2006)}]{frey2006mixture}
Frey, B.~J.; and Dueck, D. 2006.
\newblock Mixture modeling by affinity propagation.
\newblock In \emph{Advances in neural information processing systems},
  379--386.

\bibitem[{Frey and Dueck(2007)}]{frey2007clustering}
Frey, B.~J.; and Dueck, D. 2007.
\newblock Clustering by passing messages between data points.
\newblock \emph{science} 315(5814): 972--976.

\bibitem[{Glorot and Bengio(2010)}]{glorot2010understanding}
Glorot, X.; and Bengio, Y. 2010.
\newblock Understanding the difficulty of training deep feedforward neural
  networks.
\newblock In \emph{Proceedings of the thirteenth international conference on
  artificial intelligence and statistics}, 249--256.

\bibitem[{Goodfellow et~al.(2014)Goodfellow, Pouget-Abadie, Mirza, Xu,
  Warde-Farley, Ozair, Courville, and Bengio}]{goodfellow2014generative}
Goodfellow, I.; Pouget-Abadie, J.; Mirza, M.; Xu, B.; Warde-Farley, D.; Ozair,
  S.; Courville, A.; and Bengio, Y. 2014.
\newblock Generative adversarial nets.
\newblock In \emph{Advances in neural information processing systems},
  2672--2680.

\bibitem[{Gray and Scott(1984)}]{gray1984autocatalytic}
Gray, P.; and Scott, S. 1984.
\newblock Autocatalytic reactions in the isothermal, continuous stirred tank
  reactor: Oscillations and instabilities in the system A+ 2B to 3B; B to C.
\newblock \emph{Chemical Engineering Science} 39(6): 1087--1097.

\bibitem[{Gregor et~al.(2007)Gregor, Tank, Wieschaus, and
  Bialek}]{gregor2007probing}
Gregor, T.; Tank, D.~W.; Wieschaus, E.~F.; and Bialek, W. 2007.
\newblock Probing the limits to positional information.
\newblock \emph{Cell} 130(1): 153--164.

\bibitem[{Gulrajani et~al.(2017)Gulrajani, Ahmed, vsky, Dumoulin, and
  Courville}]{gulrajani2017improved}
Gulrajani, I.; Ahmed, F.; vsky, M.; Dumoulin, V.; and Courville, A.~C. 2017.
\newblock Improved training of wasserstein gans.
\newblock In \emph{Advances in neural information processing systems},
  5767--5777.

\bibitem[{Gur-Ari, Roberts, and Dyer(2018)}]{gur2018gradient}
Gur-Ari, G.; Roberts, D.~A.; and Dyer, E. 2018.
\newblock Gradient descent happens in a tiny subspace.
\newblock \emph{arXiv preprint arXiv:1812.04754} .

\bibitem[{Karacan et~al.(2016)Karacan, Akata, Erdem, and
  Erdem}]{karacan2016learning}
Karacan, L.; Akata, Z.; Erdem, A.; and Erdem, E. 2016.
\newblock Learning to generate images of outdoor scenes from attributes and
  semantic layouts.
\newblock \emph{arXiv preprint arXiv:1612.00215} .

\bibitem[{Kondo and Miura(2010)}]{kondo2010reaction}
Kondo, S.; and Miura, T. 2010.
\newblock Reaction-diffusion model as a framework for understanding biological
  pattern formation.
\newblock \emph{science} 329(5999): 1616--1620.

\bibitem[{Ledig et~al.(2017)Ledig, Theis, Husz{\'a}r, Caballero, Cunningham,
  Acosta, Aitken, Tejani, Totz, Wang et~al.}]{ledig2017photo}
Ledig, C.; Theis, L.; Husz{\'a}r, F.; Caballero, J.; Cunningham, A.; Acosta,
  A.; Aitken, A.; Tejani, A.; Totz, J.; Wang, Z.; et~al. 2017.
\newblock Photo-realistic single image super-resolution using a generative
  adversarial network.
\newblock In \emph{Proceedings of the IEEE conference on computer vision and
  pattern recognition}, 4681--4690.

\bibitem[{Lei et~al.(2019)Lei, Lee, Dimakis, and Daskalakis}]{lei2019sgd}
Lei, Q.; Lee, J.~D.; Dimakis, A.~G.; and Daskalakis, C. 2019.
\newblock Sgd learns one-layer networks in wgans.
\newblock \emph{arXiv preprint arXiv:1910.07030} .

\bibitem[{Li and Liang(2018)}]{li2018learning}
Li, Y.; and Liang, Y. 2018.
\newblock Learning overparameterized neural networks via stochastic gradient
  descent on structured data.
\newblock In \emph{Advances in Neural Information Processing Systems},
  8157--8166.

\bibitem[{Luc et~al.(2016)Luc, Couprie, Chintala, and
  Verbeek}]{luc2016semantic}
Luc, P.; Couprie, C.; Chintala, S.; and Verbeek, J. 2016.
\newblock Semantic Segmentation using Adversarial Networks.
\newblock In \emph{NIPS Workshop on Adversarial Training}.

\bibitem[{Maaten and Hinton(2008)}]{maaten2008visualizing}
Maaten, L. v.~d.; and Hinton, G. 2008.
\newblock Visualizing data using t-SNE.
\newblock \emph{Journal of machine learning research} 9(Nov): 2579--2605.

\bibitem[{Meinhardt(1982)}]{meinhardt1982models}
Meinhardt, H. 1982.
\newblock Models of biological pattern formation.
\newblock \emph{New York} 118.

\bibitem[{Meinhardt and Gierer(1974)}]{meinhardt1974applications}
Meinhardt, H.; and Gierer, A. 1974.
\newblock Applications of a theory of biological pattern formation based on
  lateral inhibition.
\newblock \emph{Journal of cell science} 15(2): 321--346.

\bibitem[{Meinhardt and Gierer(2000)}]{meinhardt2000pattern}
Meinhardt, H.; and Gierer, A. 2000.
\newblock Pattern formation by local self-activation and lateral inhibition.
\newblock \emph{Bioessays} 22(8): 753--760.

\bibitem[{Mirza and Osindero(2014)}]{mirza2014conditional}
Mirza, M.; and Osindero, S. 2014.
\newblock Conditional generative adversarial nets.
\newblock \emph{arXiv preprint arXiv:1411.1784} .

\bibitem[{Nagarajan and Kolter(2019)}]{nagarajan2019generalization}
Nagarajan, V.; and Kolter, J.~Z. 2019.
\newblock Generalization in deep networks: The role of distance from
  initialization.
\newblock \emph{arXiv preprint arXiv:1901.01672} .

\bibitem[{Nakamasu et~al.(2009)Nakamasu, Takahashi, Kanbe, and
  Kondo}]{nakamasu2009interactions}
Nakamasu, A.; Takahashi, G.; Kanbe, A.; and Kondo, S. 2009.
\newblock Interactions between zebrafish pigment cells responsible for the
  generation of Turing patterns.
\newblock \emph{Proceedings of the National Academy of Sciences} 106(21):
  8429--8434.

\bibitem[{Neyshabur et~al.(2018)Neyshabur, Li, Bhojanapalli, LeCun, and
  Srebro}]{neyshabur2018role}
Neyshabur, B.; Li, Z.; Bhojanapalli, S.; LeCun, Y.; and Srebro, N. 2018.
\newblock The role of over-parametrization in generalization of neural
  networks.
\newblock In \emph{International Conference on Learning Representations}.

\bibitem[{Rauch and Millonas(2004)}]{rauch2004role}
Rauch, E.~M.; and Millonas, M.~M. 2004.
\newblock The role of trans-membrane signal transduction in Turing-type
  cellular pattern formation.
\newblock \emph{Journal of theoretical biology} 226(4): 401--407.

\bibitem[{Rogers(2003)}]{rogers2003diffusion}
Rogers, E. 2003.
\newblock Diffusion of innovations . Delran.
\newblock \emph{NJ: Simon \& Schuster. Schneider, L.(1971). Dialectic in
  sociology. American Sociological Review} 36: 667678.

\bibitem[{Rout(2020)}]{rout2020alert}
Rout, L. 2020.
\newblock Alert: Adversarial learning with expert regularization using tikhonov
  operator for missing band reconstruction.
\newblock \emph{IEEE Transactions on Geoscience and Remote Sensing} 58(6):
  4395--4405.

\bibitem[{Rout et~al.(2020)Rout, Misra, Manthira~Moorthi, and
  Dhar}]{rout2020s2a}
Rout, L.; Misra, I.; Manthira~Moorthi, S.; and Dhar, D. 2020.
\newblock S2A: Wasserstein GAN with Spatio-Spectral Laplacian Attention for
  Multi-Spectral Band Synthesis.
\newblock In \emph{Proceedings of the IEEE/CVF Conference on Computer Vision
  and Pattern Recognition Workshops}, 188--189.

\bibitem[{Sarmad, Lee, and Kim(2019)}]{sarmad2019rl}
Sarmad, M.; Lee, H.~J.; and Kim, Y.~M. 2019.
\newblock Rl-gan-net: A reinforcement learning agent controlled gan network for
  real-time point cloud shape completion.
\newblock In \emph{Proceedings of the IEEE Conference on Computer Vision and
  Pattern Recognition}, 5898--5907.

\bibitem[{Saxe, McClelland, and Ganguli(2014)}]{saxe2014exact}
Saxe, A.; McClelland, J.; and Ganguli, S. 2014.
\newblock Exact solutions to the nonlinear dynamics of learning in deep linear
  neural networks.
\newblock International Conference on Learning Represenatations 2014.

\bibitem[{Sayama(2015)}]{sayama2015introduction}
Sayama, H. 2015.
\newblock \emph{Introduction to the modeling and analysis of complex systems}.
\newblock Open SUNY Textbooks.

\bibitem[{Tsuchida, Roosta, and Gallagher(2018)}]{tsuchida2018invariance}
Tsuchida, R.; Roosta, F.; and Gallagher, M. 2018.
\newblock Invariance of weight distributions in rectified MLPs.
\newblock In \emph{International Conference on Machine Learning}, 4995--5004.

\bibitem[{Turing(1952)}]{turing1952chemical}
Turing, A. 1952.
\newblock THE CHEMICAL BASIS OF MORPHOGENESIS.
\newblock \emph{Philosophical Transactions of the Royal Society of London.
  Series B, Biological Sciences} 237(641): 37--72.

\bibitem[{Verhulst(1838)}]{verhulst1838notice}
Verhulst, P.-F. 1838.
\newblock Notice sur la loi que la population suit dans son accroissement.
\newblock \emph{Corresp. Math. Phys.} 10: 113--126.

\bibitem[{Vershynin(2018)}]{vershynin2018high}
Vershynin, R. 2018.
\newblock \emph{High-dimensional probability: An introduction with applications
  in data science}, volume~47.
\newblock Cambridge university press.

\bibitem[{Wang and Gupta(2016)}]{wang2016generative}
Wang, X.; and Gupta, A. 2016.
\newblock Generative image modeling using style and structure adversarial
  networks.
\newblock In \emph{European conference on computer vision}, 318--335. Springer.

\bibitem[{Wang et~al.(2018)Wang, Yu, Wu, Gu, Liu, Dong, Qiao, and
  Change~Loy}]{wang2018esrgan}
Wang, X.; Yu, K.; Wu, S.; Gu, J.; Liu, Y.; Dong, C.; Qiao, Y.; and Change~Loy,
  C. 2018.
\newblock Esrgan: Enhanced super-resolution generative adversarial networks.
\newblock In \emph{Proceedings of the European Conference on Computer Vision
  (ECCV)}, 0--0.

\bibitem[{Wolpert, Tickle, and Arias(2015)}]{wolpert2015principles}
Wolpert, L.; Tickle, C.; and Arias, A.~M. 2015.
\newblock \emph{Principles of development}.
\newblock Oxford University Press, USA.

\bibitem[{Xie, Liang, and Song(2017)}]{xie2017diverse}
Xie, B.; Liang, Y.; and Song, L. 2017.
\newblock Diverse neural network learns true target functions.
\newblock In \emph{Artificial Intelligence and Statistics}, 1216--1224.

\bibitem[{Zhu et~al.(2017)Zhu, Park, Isola, and Efros}]{zhu2017unpaired}
Zhu, J.-Y.; Park, T.; Isola, P.; and Efros, A.~A. 2017.
\newblock Unpaired image-to-image translation using cycle-consistent
  adversarial networks.
\newblock In \emph{Proceedings of the IEEE international conference on computer
  vision}, 2223--2232.

\end{thebibliography}
	
	\clearpage
	
	\appendix
	\section{Appendix}
	\section{More Related Works}
	\label{rw}
	\subsection{Reaction-Diffusion Systems}
	The original RD model is a simplification and an idealization of practical systems whose complexity makes it hard to understand the phenomena. With slight modification to the theory, one may easily extend this mathematical analysis to explain pattern formation in real world systems. In addition, it can generate limitless variety of patterns depending on the parameters in the reaction and diffusion terms.
	
	Numerous methods seek to explain pattern formation in complex systems. Among many reasonable attempts, one that experimental biologists may recall is gradient model~\cite{wolpert2015principles}. Different from RD model, it assumes a fixed source of morphogens that provides positional information. In other words, it can be designed as a special case of RD model by carefully choosing the boundary conditions. Experiments have shown the necessity of molecular interaction and boundary condition to create more realistic patterns~\cite{meinhardt1982models}. To model interactions of molecular elements in gradient analysis, \cite{gregor2007probing} developed a framework that is essentially similar to RD model.
	
	Concerted efforts have been made towards extension and identification of root causes to explain pattern formation. The fact that a short range positive feedback and a long range negative feedback are enough to generate Turing patterns is indeed a big revelation in this direction~\cite{meinhardt1974applications,meinhardt2000pattern}. This refinement helps envision a wide variety of patterns in more complex systems. 
	
	Particularly intriguing is the fact that these interacting elements need not be limited to molecules. The interaction between cellular signals also generates Turing patterns~\cite{nakamasu2009interactions}. Further, there is no restriction on how the system diffuses to break spatial symmetry. A relayed series of cell to cell signal transmission may induce diffusible factors in a system~\cite{rauch2004role}. All these scenarios have a common ground in a sense that these systems exhibit a short range positive feedback and a long range negative feedback similar to adversarial framework.
	
	\subsection{Bernoulli Differential Equation}
	Bernoulli differential equation was discussed by \citeauthor{bernoulli1695explicationes} in \citeyear{bernoulli1695explicationes}. This fundamental equation arises naturally in a wide variety fields, such as modelling of population growth~\cite{verhulst1838notice}, modelling of a pandemic, modelling of growth of tumors, Fermi-Dirac statistics, modelling of crop response, and modelling of diffusion of innovations in economics and sociology~\cite{rogers2003diffusion}. In \citeauthor{verhulst1838notice} model of population growth, the rate of reproduction is proportional to current population and available resources. Formally,
	\begin{equation}
	\begin{split}
	\frac{d P}{d t} = rP\left ( 1 - \frac{P}{K}\right ),
	\end{split}
	\end{equation}
	where $P, r,$ and $K$ denote population size, rate of growth, and carrying capacity, respectively. In ecology, $N$ is often used in place of $P$ to represent population. An interesting theory, namely $r/K$ selection theory builds on simplified \citeauthor{verhulst1838notice} model by drawing $r$ and $K$ from ecological algebra. In the similar spirit, the presented PRD model arrives at a special Bernoulli differential equation where error being the population size in the modelling of population growth. Different from traditional settings, the rate of change here is proportional to the square root of current error. To put more succinctly, 
	\begin{equation}
	\begin{split}
	\frac{d \psi}{d t} \leq r \psi^{1/2}\left ( 1 - \frac{\psi^{1/2}}{K}\right ), 
	\end{split}
	\end{equation}
	where $r = \lambda_1 \mu$ and $K = \kappa \mu$. The interpretation of this equation is reversed in the present analysis as we are interested in the decay of total error. Nevertheless, the compact representation captures the essence of reaction and diffusion dynamics.

	\section{Jointly Training Both Layers}
	\label{both_layers}
	In this section, we extend theoretical analyses from a single layer scalar network architecture to jointly training both layers with multiple classes. For simplicity, let $\vz_p$ denotes $\frac{1}{\sqrt{d_{out}m}} \mV \sigma\left ( \mU \vx_p \right )$.
	
	\textbf{Definition 2.} \textit{Let us define $\mathfrak{R}^{\vu}_{j}\left ( \vu_j, \vv_j \right )$ and $\mathfrak{R}^{\vv}_{j}\left ( \vu_j, \vv_j \right )$ as the reaction terms in hidden and top layer, respectively. Also, let $\mathfrak{D}^{\vu}_{j}\left ( \nabla^2 \vu_j \right )$ and $\mathfrak{D}^{\vv}_{j}\left ( \nabla^2 \vv_j \right )$ denote the diffusion terms in hidden and top layer, respectively.}

	\textbf{Theorem 3.} (Reaction-Diffusion Dynamics) \textit{ If we absorb constants in $\mathcal{O}(.)$ and set $\left ( \vy_p - \vz_p \right )_i  v_{ij} 1_{\left \{  \vu_j^T\vx_p  \geq 0 \right \}} x_{p,k} = \mathcal{O}\left ( 1 \right )$ for $i \in [d_{out}]$ and $p \in [n]$, then for all $j\in[m]$ the RD dynamics satisfy:
		\begin{equation*}
		\begin{split}
		\mathfrak{R}^{\vu}_{j}\left ( \vu_j, \vv_j \right ) & =\mathcal{O}\left (n d_{in}\sqrt{ \frac{d_{out}}{m}} \right ),\\
		\mathfrak{D}^{\vu}_{j}\left ( \nabla^2 \vu_j \right ) & = \mathcal{O}\left (nm^2 d_{in} d_{out}^{3/2}\right ),\\
		\mathfrak{R}^{\vv}_{j}\left ( \vu_j, \vv_j \right ) & = \mathcal{O}\left ( n d_{in} \sqrt{\frac{d_{out}}{m}} \right ),\\
		\mathfrak{D}^{\vv}_{j}\left ( \nabla^2 \vv_j \right ) & = \mathcal{O}\left (nm^2 d_{in} d_{out}^{1/2}\right ).
		\end{split}
		\end{equation*}}
	\textit{Proof.} See next section. The diffusion terms are greatly affected by other morphogens in the system, suggesting a special case scenario of Turing's RD model. To put more succinctly, $\mathfrak{D}^{\vu}_{j}$ and $\mathfrak{D}^{\vv}_{j}$ are dominated by $\vv_j$ and $\vu_j$, respectively. While the asymptotic reaction terms are bounded by similar norms, the apparent difference between diffusion terms explains why we observe different patterns in the hidden and top layer.

	\section{Technical Proofs}
	\label{proofs}
	
	\subsection{Proof of Lemma 1}
	\label{pr_lm1}
	
	With probability at least $(1-\delta)$, by Markov's inequality, we get
	\begin{equation}
	\left | u_{jk}\left ( 0 \right ) \right |\leq \frac{\mathbb{E}\left [ \left | u_{jk}\left ( 0 \right ) \right | \right ]}{\delta} = \frac{2}{\sqrt{2\pi} \delta}.
	\end{equation}
	We use matrix norm properties to bound the Frobenius norm of $U(0)$:
	\begin{equation}
	\begin{split}
	\left \| U(0) \right \|_F &= \left ( \sum_{j=1}^{m} \sum_{k=1}^{d_{in}} \left | u_{jk}(0) \right |^2 \right )^{1/2}\\
	&\leq \left ( \sum_{j=1}^{m} \sum_{k=1}^{d_{in}} \frac{4}{2\pi \delta^2} \right )^{1/2}\\
	&\leq \frac{2\sqrt{m d_{in}}}{\sqrt{2\pi}\delta} \triangleq \zeta.
	\end{split}
	\end{equation}
	This finishes the proof of the statement.	\hfill $\square$
	
	\subsection{Proof of Lemma 4}
	For clarity, let us recall \textbf{Lemma 3.1} of \citeauthor{du2018gradient}: \textit{If $m = \Omega\left \{ \frac{n^2}{\lambda_0^2} \log\left ( \frac{n}{\delta} \right)\right \}$, then we have with high probability $1-\delta$, $\left \| \mathcal{H}(0) - \mathcal{H}^{\infty} \right \|_2 \leq \frac{\lambda_0}{4}$ and $\lambda_{\min}\left (\mathcal{H}(0) \right) \geq \frac{3}{4}\lambda_0$.} From \textbf{Remark 1}, we know 
	\begin{equation}
	\begin{split}
	\left \| \mathcal{H}\right \|_2 - \left \|\mathcal{H}(0) \right \|_2 \leq \left \| \mathcal{H} - \mathcal{H}(0) \right \|_2 \leq \frac{\lambda_0}{4}.
	\end{split}
	\end{equation}
	Using similar arguments, we get 
	\begin{equation}
	\begin{split}
	\left \| \mathcal{H}(0)\right \|_2 - \left \|\mathcal{H}^{\infty} \right \|_2 \leq \left \| \mathcal{H}(0) - \mathcal{H}^{\infty} \right \|_2 \leq \frac{\lambda_0}{4},
	\end{split}
	\end{equation}
	which implies $\lambda_{\max}(\mathcal{H}(0)) \leq \lambda_{\max}(\mathcal{H}^{\infty})+\frac{\lambda_0}{4}$. By plugging this, the expression gets simplified to 
	\begin{equation}
	\begin{split}
	\lambda_{\max}\left( \mathcal{H} \right)  &\leq \lambda_{\max}\left(\mathcal{H}^{\infty} \right) + \frac{\lambda_0}{4} + \frac{\lambda_0}{4}\\
	& \leq \lambda_1^{\infty} + \frac{\lambda_0}{2} \triangleq \frac{\lambda_1}{2}.
	\end{split}
	\end{equation}
	This justifies the upper bound assumption of the largest eigen value over iterations.\hfill$\square$

	\subsection{Proof of Theorem 3}
	\label{pr_th3}
	Our first step is to derive dynamics of weights due to supervised cost. In the hidden layer, the weights are updated by the following PDE.
	
	\begin{equation}
	\begin{split}
	&\frac{\partial \mathcal{L}_{sup}\left ( \mU, \mV \right )}{\partial u_{jk}} \\&= \frac{1}{2} \sum_{p=1}^{n} \frac{\partial}{\partial u_{jk}} \sum_{i=1}^{d_{out}}\left ( \vz_p - \vy_p \right )_i^{2}\\
	&= \sum_{p=1}^{n}\sum_{i=1}^{d_{out}} \left ( \vz_p -\vy_p \right )_i \frac{\partial z_{p,i}}{\partial u_{jk}}\\
	&= \sum_{p=1}^{n}\sum_{i=1}^{d_{out}} \left ( \vz_p - \vy_p \right )_i \frac{1}{\sqrt{d_{out}m}} v_{ij}\frac{\partial}{\partial u_{jk}} \sigma\left ( \vu_j^T\vx_p  \right )\\
	&=\frac{1}{\sqrt{d_{out}m}}\sum_{p=1}^{n}\sum_{i=1}^{d_{out}} \left ( \vz_p - \vy_p \right )_i  v_{ij} 1_{\left \{  \vu_j^T\vx_p  \geq 0 \right \}} x_{p,k}.
	\end{split}
	\end{equation}
	
	Next, we calculate dynamics of weights in the top layer.
	\begin{equation}
	\begin{split}
	&\frac{\partial \mathcal{L}_{sup}\left ( \mU, \mV \right )}{\partial v_{ij}} \\& = \frac{1}{2} \sum_{p=1}^{n} \frac{\partial}{\partial v_{ij}} \sum_{i=1}^{d_{out}}\left ( \vz_p - \vy_p \right )_i^{2}\\
	&= \sum_{p=1}^{n}  \left ( \vz_p - \vy_p \right )_i \frac{\partial  z_{p,i}}{\partial v_{ij}}\\
	&= \sum_{p=1}^{n}  \left ( \vz_p - \vy_p \right )_i \frac{1}{\sqrt{d_{out}m}} \frac{\partial }{\partial v_{ij}} \sum_{j=1}^{m} v_{ij} \sigma\left (  \vu_j^T \vx_p  \right )\\
	&= \frac{1}{\sqrt{d_{out}m}} \sum_{p=1}^{n}  \left ( \vz_p - \vy_p \right )_i  1_{\left \{  \vu_j^T \vx_p  \geq 0 \right \}}  \vu_j^T \vx_p .
	\end{split}
	\end{equation}
	
	Now, we proceed to compute dynamics of weights due to adversarial regularization. In the hidden layer, the weights obey the following dynamics:
	\begin{equation}
	\begin{split}
	&\frac{\partial \mathcal{L}_{adv}\left ( \mU, \mV, \mW, \va \right )}{\partial u_{jk}} \\&= \frac{1}{m\sqrt{d_{out}}} \sum_{p=1}^{n}  \va^T \text{diag}\left ( 1_{\left \{ \mW \mV \sigma\left ( \mU \vx_p \right ) \geq 0 \right \}}  \right ) \mW \vv_j 1_{\left \{  \vu_j^T \vx_p  \geq 0 \right \}} x_{p,k}.
	\end{split}
	\end{equation}
	
	The weights in the top layer are governed by:
	\begin{equation}
	\begin{split}
	&\frac{\partial \mathcal{L}_{adv}\left ( \mU, \mV, \mW, \va  \right )}{\partial v_{ij}} \\&= \frac{1}{m\sqrt{d_{out}}} \sum_{p=1}^{n}  \va^T \text{diag}\left ( 1_{\left \{ \mW \mV \sigma\left ( \mU \vx_p \right ) \geq 0 \right \}}  \right ) \mW_{:,i} 1_{\left \{  \vu_j^T \vx_p  \geq 0 \right \}}  \vu_j^T \vx_p .
	\end{split}
	\end{equation}
	
	Analogous to equation~(\ref{rd-eq}), the reaction and diffusion terms in augmented objective are defined as:
	\begin{equation}
	\begin{split}
	&\mathfrak{R}^{\vu}_{j}\left ( \vu_j, \vv_j \right ) \\&\triangleq \left \{\frac{1}{\sqrt{d_{out}m}}\sum_{p=1}^{n}\sum_{i=1}^{d_{out}} \left ( \vy_p - \vz_p \right )_i  v_{ij} 1_{\left \{  \vu_j^T\vx_p  \geq 0 \right \}} x_{p,k} \right \}_{k=1}^{d_{in}},
	\end{split}
	\label{prd-eq1}
	\end{equation}

	\begin{equation}
	\begin{split}
	&\mathfrak{D}^{\vu}_{j}\left ( \nabla^2 \vu_j \right ) \\&\triangleq \left \{\frac{1}{m\sqrt{d_{out}}} \sum_{p=1}^{n}  \va^T \text{diag}\left ( 1_{\left \{ \mW \mV \sigma\left ( \mU \vx_p \right ) \geq 0 \right \}}  \right ) \mW \vv_j 1_{\left \{  \vu_j^T \vx_p  \geq 0 \right \}} x_{p,k} \right \}_{k=1}^{d_{in}},\\
	\end{split}
	\label{prd-eq2}
	\end{equation}
	
	\begin{equation}
	\begin{split}
	&\mathfrak{R}^{\vv}_{j}\left ( \vu_j, \vv_j \right ) \\&\triangleq  \left \{\frac{1}{\sqrt{d_{out}m}} \sum_{p=1}^{n}  \left ( \vy_p - \vz_p \right )_i  1_{\left \{  \vu_j^T \vx_p  \geq 0 \right \}}  \vu_j^T \vx_p \right \}_{i=1}^{d_{out}},\\
	\end{split}
	\label{prd-eq3}
	\end{equation}
	
	\begin{equation}
	\begin{split}
	&\mathfrak{D}^{\vv}_{j}\left ( \nabla^2 \vv_j \right ) \\&\triangleq  \left \{  \frac{1}{m\sqrt{d_{out}}} \sum_{p=1}^{n}  \va^T \text{diag}\left ( 1_{\left \{ \mW \mV \sigma\left ( \mU \vx_p \right ) \geq 0 \right \}}  \right ) \mW_{:,i} 1_{\left \{  \vu_j^T \vx_p  \geq 0 \right \}}  \vu_j^T \vx_p \right \}_{i=1}^{d_{out}}. \\
	\end{split}
	\label{prd-eq4}
	\end{equation}

	Ignoring constants and assuming $\left ( \vy_p - \vz_p \right )_i  v_{ij} 1_{\left \{  \vu_j^T\vx_p  \geq 0 \right \}} x_{p,k} = \mathcal{O}\left ( 1 \right )$, we get asymptotic bounds on the norm of reaction and diffusion terms\footnote{More precisely, one may choose a generator to have different number of hidden units than discriminator. In that case, the asymptotic bounds may contain $m_{dis}$ and $m_{gen}$. To simplify the expression and focus more on analysis, we assume equal number of hidden units in generator and discriminator.}:
	\begin{equation}
	\begin{split}
	\mathfrak{R}^{\vu}_{j}\left ( \vu_j, \vv_j \right ) & =\mathcal{O}\left (n d_{in}\sqrt{ \frac{d_{out}}{m}} \right ),\\
	\mathfrak{D}^{\vu}_{j}\left ( \nabla^2 \vu_j \right ) & = \mathcal{O}\left (nm^2 d_{in} d_{out}^{3/2}\right ),\\
	\mathfrak{R}^{\vv}_{j}\left ( \vu_j, \vv_j \right ) & = \mathcal{O}\left ( n d_{in} \sqrt{\frac{d_{out}}{m}} \right ),\\
	\mathfrak{D}^{\vv}_{j}\left ( \nabla^2 \vv_j \right ) & = \mathcal{O}\left (nm^2 d_{in} d_{out}^{1/2}\right ),
	\end{split}
	\end{equation}
	which completes the proof.\hfill $\square$

	\section{Experiments}
	\label{exps}
	Our experiments aim to answer the following questions:
	\begin{itemize}
		\item (\textbf{Linear Rate}) \textit{How does randomly initialized gradient descent converge exponentially fast to a solution in a larger subspace around initialization?}
		\item (\textbf{Theorem 1}) \textit{How does purely supervised cost maintain symmetry and homogeneity over iterations?}
		\item (\textbf{Theorem 2}) \textit{How does adversarial interaction help break this symmetry and homogeneity? }
	\end{itemize}

	\subsection{Training Details} 
	\subsubsection{Architecture} We use two layer neural network with ReLU activation in all the experiments. Refer to section \textbf{Preliminaries} for mathematical details on the network architecture. The discriminator is updated once per each generator update. In addition to gradient penalty, we set $L=0.01$ for the discriminator. The primary results are reported with the number of hidden units, $m=2^{13}$ to capture over-parameterization. Also, we experiment with 13 different hidden units ranging from $m=2^{3}$ to $2^{15}$ so as to study the generalization of our theory. The input to the network is $d_{in}=784$ on synthetic dataset, MNIST and FashionMNIST. We also study the effect of $d_{in}=1024$, which is like gray scale CIFAR10 dataset, on pattern formation. The output of the network is set to $d_{out}=10$ in all experiments. The specific choice of scalar network, i.e., $d_{out}=1$ is made for mathematical convenience though we demonstrate empirically that the theory can be successfully applied to analyze more complex networks in less restrictive setting. A system with 64GB RAM, one V100 gpu and PyTorch library is used for all the experiments reported in this paper.
	
	\subsubsection{Synthetic Datasets} 
\begin{figure}[t]
		\centering
		\includegraphics[width=\linewidth]{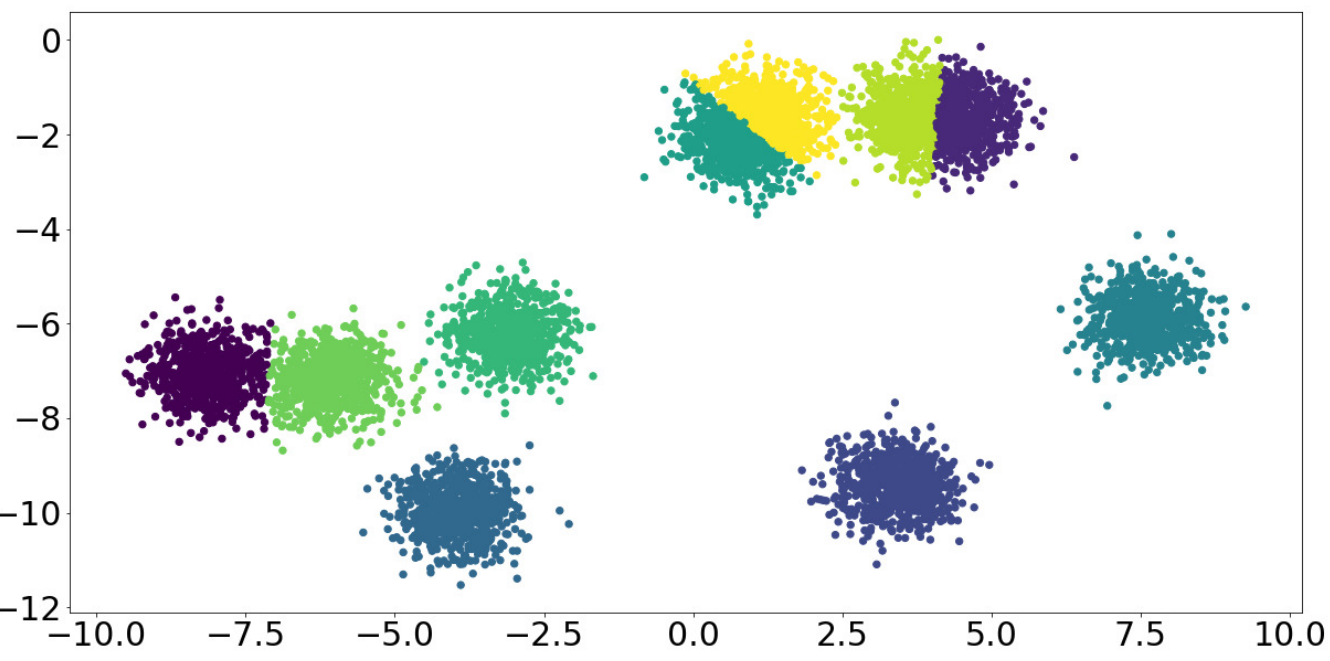}
		\caption{Synthetic dataset on 2-dimensional manifold.}
		\label{synthetic}
	\end{figure}

It has been established in multitude of tasks that natural images often lie on a low dimensional manifold. To emulate this structure, we have created a synthetic dataset: $\left \{\left ( \vx_p, \vy_p \right )  \right \}_{p=1}^{n} \subset \mathbb{R}^{784} \times \mathbb{R}^{10}$. To enforce data points to lie on a low dimensional manifold (here, 2), the initial two elements of $\vx_p$ are sampled from a mixture of Gaussian distribution having $d_{in}=10$ different modes similar to MNIST, FashionMNIST and CIFAR10. The rest of the elements of $\vx_p$ are set to 1. In another dataset, we set $d_{in}=1024$ with initial two elements sampled from the same mixture of Gaussian distribution. In both these datasets, we have $n=7000$ samples with $6000$ used for training and remaining $1000$ used for testing. We use a batch size of 256, a learning rate of $1e-2$, and train for 1000 epochs using SGD optimizer with momentum 0.9. The random seed is set to 1 in the mixture of Gaussian distribution (ref. \verb|make_blobs| from \verb|sklearn|). Figure~\ref{synthetic} illustrates different classes of this toy dataset.
	
	\subsubsection{MNIST and FashionMNIST Datasets} Here, out of $n=70000$ samples of size [$28 \times 28$], we select $60000$ for training and $10000$ for testing purpose. We use a batch size of 512, a learning rate of $1e-2$, and train for 1000 epochs or until convergence using SGD optimizer with momentum 0.9. In this setting, $d_{in}=784$. We assume convergence when the training error reaches at least 0.001. 
	
	\subsubsection{CIFAR10 Dataset} Out of $60000$ samples of size [$32\times32\times 3$], we select 50000 for training and remaining 10000 for testing. Here, $d_{in}$ is set to 3072. Other than the convergence criterion which is set to 0.02, all the hyperparameters are identical to MNIST and FashionMNIST.


	\subsection{Experimental Results} 
\begin{figure}[t]
		\centering
		\includegraphics[width=0.9\linewidth]{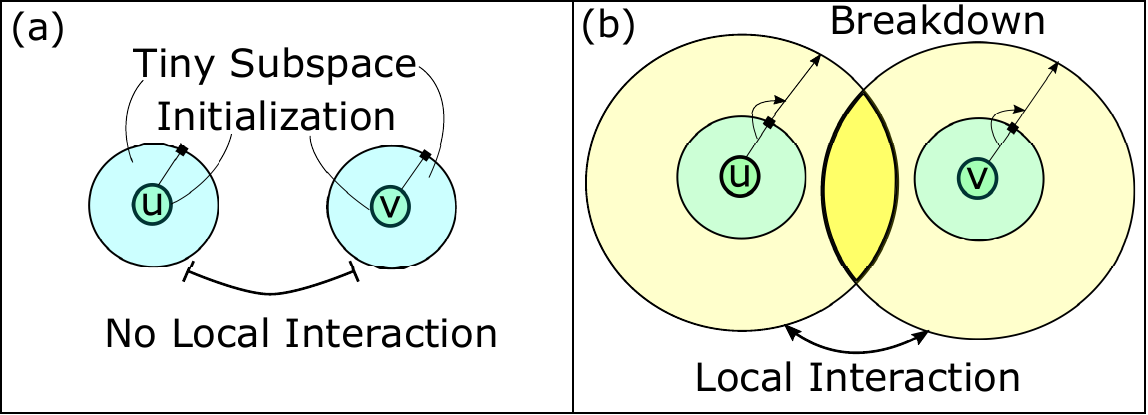}
		\caption{Breakdown of symmetry and homogeneity leads to local interaction. (a) Without Diffusion. (b) With Diffusion. }
		\label{overview}
	\end{figure}

	The empirical evidence is outlined as following. First, we demonstrate that the spatial grid of neurons follows an exponential upper bound as predicted by our theory. Second, we show that the system of neurons in sole supervision preserves symmetry and homogeneity over iterations as per \textbf{Theorem 1}. Finally, we verify \textbf{Theorem 2} by showcasing Turing-like patterns under the influence of adversarial interaction. More importantly, these computer simulations on toy datasets resemble Turing-like patterns formed while training on real datasets, suggesting both share similar underlying principles. An artistic view of the breakdown of symmetry and how it leads to pattern formation as a result of local interaction is shown in Figure~\ref{overview}.
	
	\subsubsection{Results on Synthetic Datasets}
\begin{figure}[t]
		\centering
		\includegraphics[width=\linewidth]{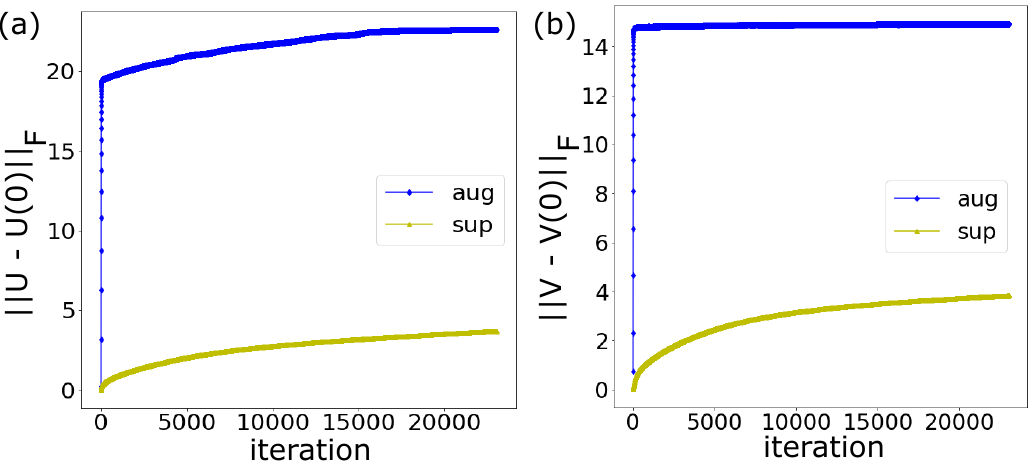}
		\caption{Distance from initialization in the (a) hidden layer (b) top layer over iterations on MNIST-like ($d_{in} = 784$) toy dataset.}
		\label{syn12_full_sup_aug_part_dis_init}
	\end{figure}

\begin{figure}[t]
		\centering
		\includegraphics[width=\linewidth]{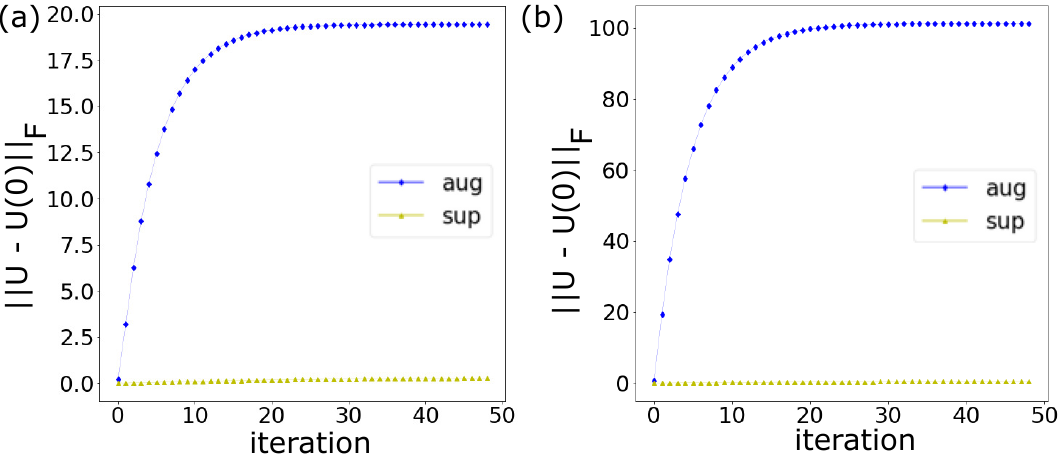}
		\caption{Distance from initialization in the \textit{hidden layer} on (a) MNIST-like ($d_{in} = 784$) and (b) CIFAR10-like ($d_{in} = 1024$) toy datasets.}
		\label{syn12_sup_aug_part_dis_init}
	\end{figure}

\begin{figure}[t]
		\centering
		\includegraphics[width=\linewidth]{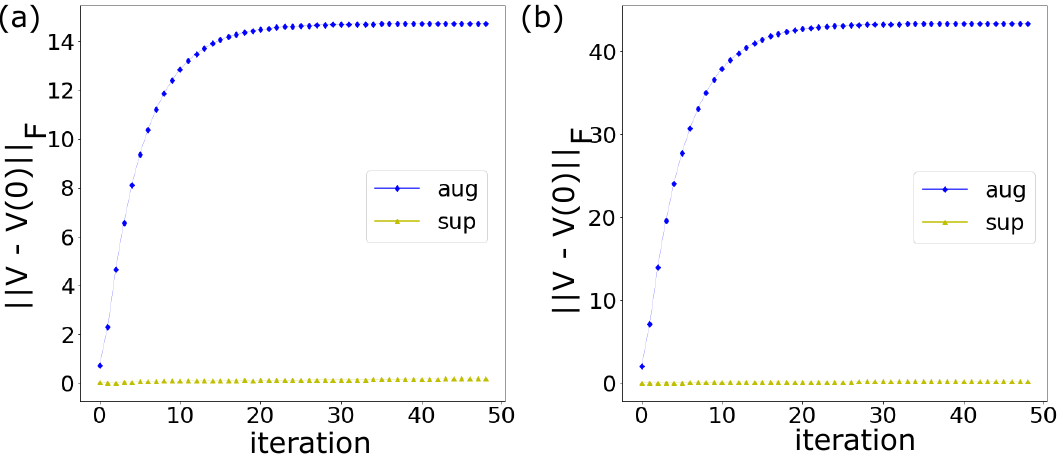}
		\caption{Distance from initialization in the \textit{top layer} on (a) MNIST-like ($d_{in} = 784$) and (b) CIFAR10-like ($d_{in} = 1024$) toy datasets.}
		\label{syn12_sup_aug_v_part_dis_init}
	\end{figure}

	While the experiments suggest approximately 2300 iterations are enough to achieve an acceptable test error (here, $\epsilon = 0.016$) on this toy dataset, we continued training up to 23000 iterations to study the impact of $\mu t$ on distance from initialization. As shown in Figure~\ref{syn12_full_sup_aug_part_dis_init}, the upper bound is dominated by the magnitude of exponential term, i.e., $\mathcal{O}\left ( \frac{n^{3/2}}{\lambda_0\delta} \right )$. Observe that $\mu$ is an infinitesimally small quantity in over-parameterized network which justifies tiny linear slope. In fact, diffusion coefficients are often orders of magnitude smaller than reaction coefficents~\cite{turing1952chemical,gray1984autocatalytic}\footnote{The reader is referred to the simulated patterns by RD model and Gray-Scott model in later parts of this paper where the impact of tiny diffusibility is discernible.}. Unless the model is trained for infinite iterations, it is fair to assume the settling point of the upper bound to be $\mathcal{O}\left ( \frac{n^{3/2}}{\lambda_0 \delta} \right )$ with a very small tolerance of $\mathcal{O}\left ( \mu T_0  \right )$. To further our understanding of exponential rate, we analyze initial few iterations on MNIST- and CIFAR10-like toy datasets. Figure~\ref{syn12_sup_aug_part_dis_init} and \ref{syn12_sup_aug_v_part_dis_init} support our theoretical analysis as we observe that $\left \| \mU(t) - \mU(0) \right \|_F$ and $\left \| \mV(t) - \mV(0) \right \|_F$ are of the form $\alpha \left (1 - \exp(-\beta t) \right) $ for $\alpha >0$ and $\beta > 0$.

	\subsubsection{Results on MNIST} 
\begin{figure}[t]
		\centering
		\includegraphics[width=\linewidth]{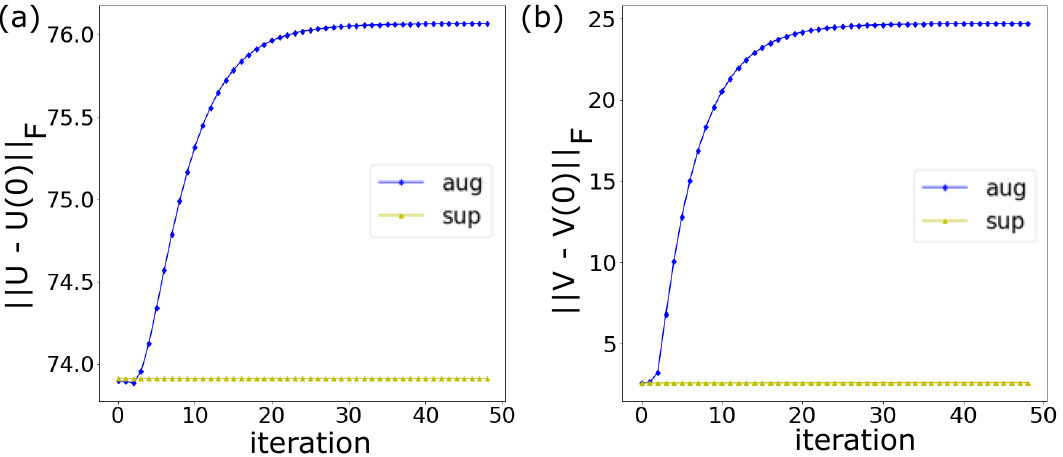}
		\caption{Distance from initialization in the (a) hidden layer (b) top layer on MNIST.}
		\label{mnist_sup_aug_part_dis_init}
	\end{figure}

\begin{figure}[t]
		\centering
		\includegraphics[width=\linewidth]{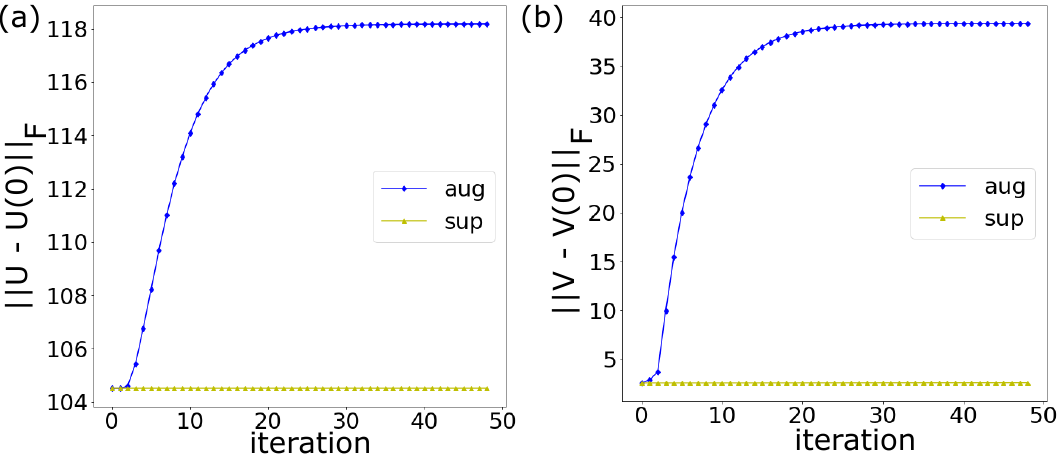}
		\caption{Distance from initialization in the (a) hidden layer (b) top layer on MNIST with $2^{14}$ hidden units.}
		\label{mnistw_sup_aug_part_dis_init}
	\end{figure}

\begin{figure}[t]
		\centering
		\includegraphics[width=\linewidth]{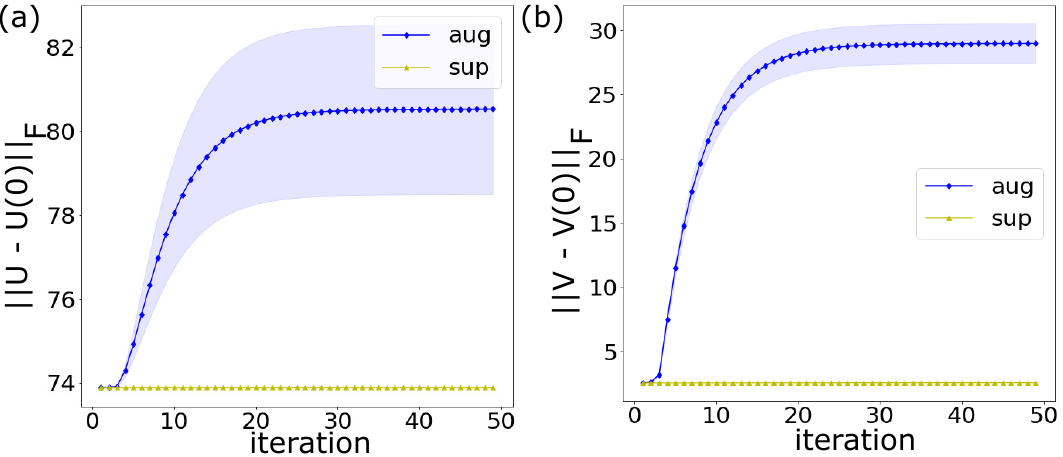}
		\caption{Distance from multiple initialization in the (a) hidden layer and (b) top layer on MNIST. }
		\label{mnist_init_sup_aug_part_dis_init}
	\end{figure}

As shown in Figure~\ref{mnist_sup_aug_part_dis_init}, we observe breakdown of symmetry to explore larger subspace on MNIST. In addition, we study the dependency of these phenomena on width in  Figure~\ref{mnistw_sup_aug_part_dis_init} and random initialization in Figure~\ref{mnist_init_sup_aug_part_dis_init} to gain a better understanding. 
	
	\textbf{Dependence on Width.} While the settling point is different in $m=2^{14}$ as compared to $m=2^{13}$, it still provides empirical support to exponential breakdown due to adversarial interaction.
	
	\textbf{Dependence on Initialization.} To investigate robustness, it is essential to run the experiments multiple times with random seeds. For this reason, we run the experiments 5 times each for supervised and adversarial learning, i.e., without and with diffusible factors. In adversarial setting, the PRD model is shown to break the symmetric and homogeneous barrier exponentially fast. Further, the experiments are consistent over multiple initializations. Though the variance appears to be quite large in hidden layer, it is relatively small ($< 2.5\%$) compared to the settling point in each trajectory. Nevertheless, it agrees with exponential rate in all the experiments.
	
\begin{figure}[!h]
		\centering
		\includegraphics[width=\linewidth]{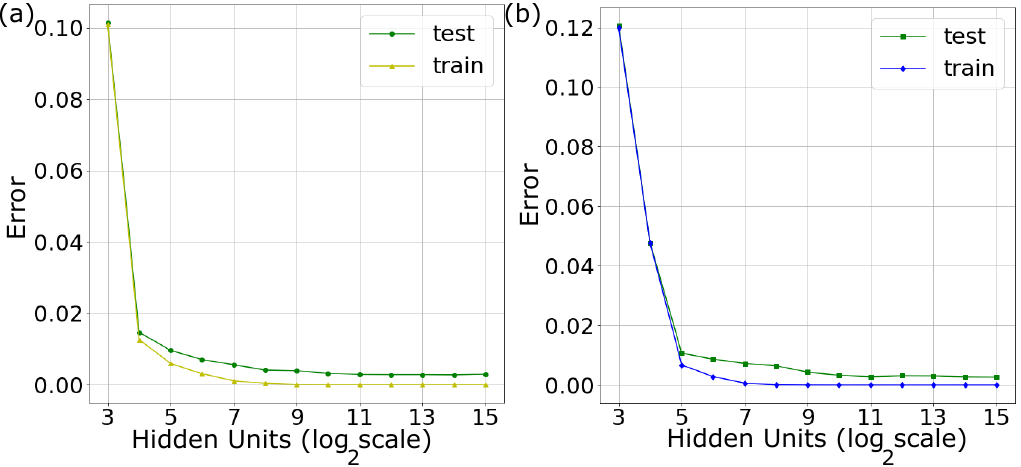}
		\caption{Training and testing error on MNIST. (a) Without Diffusion. (b) With Diffusion. }
		\label{mnist_sup_aug_err}
	\end{figure}

	\textbf{Error Analysis.} It is well known that two layer neural networks with over-parameterization can find an $\epsilon$-stationary point on MNIST in a finite time~\cite{neyshabur2018role,nagarajan2019generalization}. For clarity purpose, we conduct experiments in this problem setting. Figure~\ref{mnist_sup_aug_err} provides further evidence of this phenomena and admits the theoretical results.
	
\begin{figure*}[!h]
		\centering
		\includegraphics[width=\linewidth]{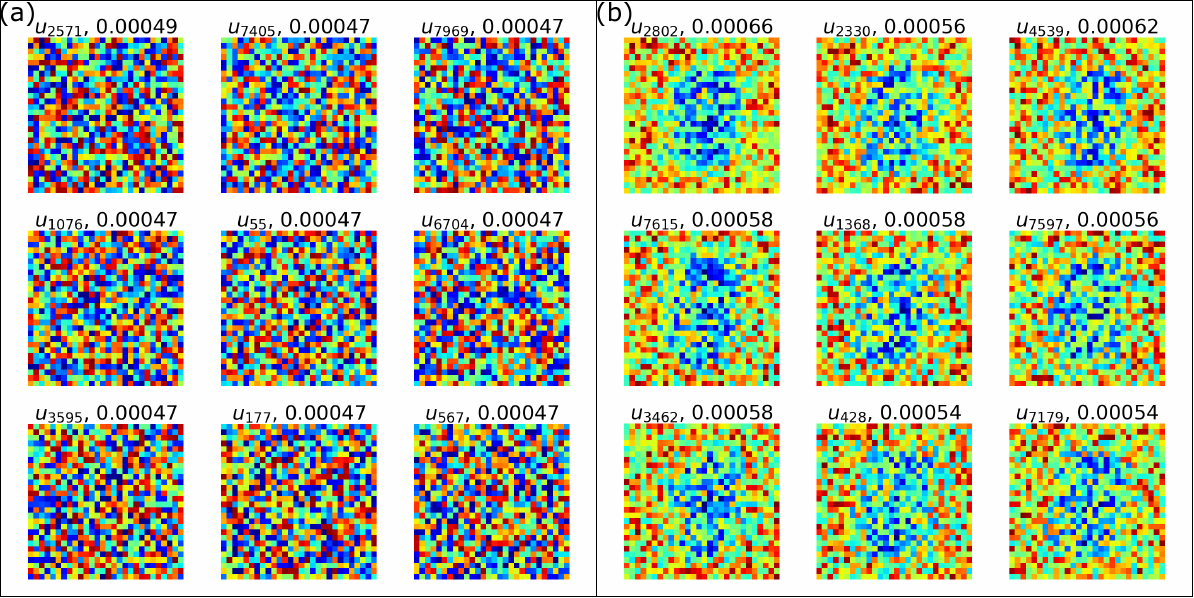}
		\caption{Comparing hidden layer filters on MNIST. For $j \in [2^{13}]$, $\left \|\vu_j \right \|_2$ is in descending order from left to right and top to bottom. (a) Without Diffusion. (b) With Diffusion. Filters with diffusion have high variances compared to without diffusion. Since digits are centred in MNIST, adversarial interaction learns to discriminate background and foreground information. }
		\label{mnist_sup_aug_ker_var}
	\end{figure*}

	\textbf{Weight Analysis.} In Figure~\ref{mnist_sup_aug_ker_var}, we plot hidden layer filters, $\vu_j$  with top-9 $L_2$ norm. An interesting observation is the organization of weights in the spatial grid of size [$28\times 28$]. Since the breakdown of symmetry allows more local interaction, the neurons coalesce in the decision making process. As shown in Figure~\ref{mnist_sup_aug_ker_var}, neurons in the hidden layer learn to focus on relevant parts of the input image in adversarial setting. This is in contrast to supervised setting where neurons pay attention to all spatial locations. To quantitatively analyze the non-homogeneity of a single node, the variance is shown alongside the hidden unit. The high variance in advesarial setting captures non-homogeneity up to some extent.
	
	\textbf{Feature Visualization.} Here, we try to understand the sensitivity of neurons to perturbations in the input space. For this reason, a common random input image is fed as input to both the systems without and with diffusion. The output of a particular neuron in the hidden layer is maximized subject to perturbations in the input image~\cite{allen2020backward}. Formally, for all $j \in[2^{13}]$,
	\begin{equation}
	\delta_j = \arg \max_{\delta \in \Delta} \vu_j^T\left (x+\delta \right),
	\end{equation} 
	where $x$ denotes a random vector in $\mathbb{R}^{d_{in}}$, $\delta$ denotes the perturbation, and $\Delta$ is the allowed set of perturbation. While one can explore different choices of allowed pertubation, a common choice, which we use in feature visualization, is $l_\infty$ ball:
	\begin{equation}
	\Delta = \left \{ \delta : \left \|\delta \right \|_\infty \leq \epsilon \right \}
	\end{equation}
	The $l_\infty$ norm of a vector in $\mathbb{R}^{d_{in}}$ is defined as:
	\begin{equation}
	\left \|\delta \right \|_\infty = \max_k \left | \delta_k\right |,
	\end{equation}
	where $k \in [d_{in}]$. Basically, the perturbation in each component is restricted to $[-\epsilon, \epsilon]$. Here, we choose $\epsilon=0.007$, and run gradient descent with learning rate $1e-1$ for $100$ iterations. The extracted features are reshaped to $[28\times 28]$ for visualization purpose.
	
\begin{figure*}[!h]
		\centering
		\includegraphics[width=\linewidth]{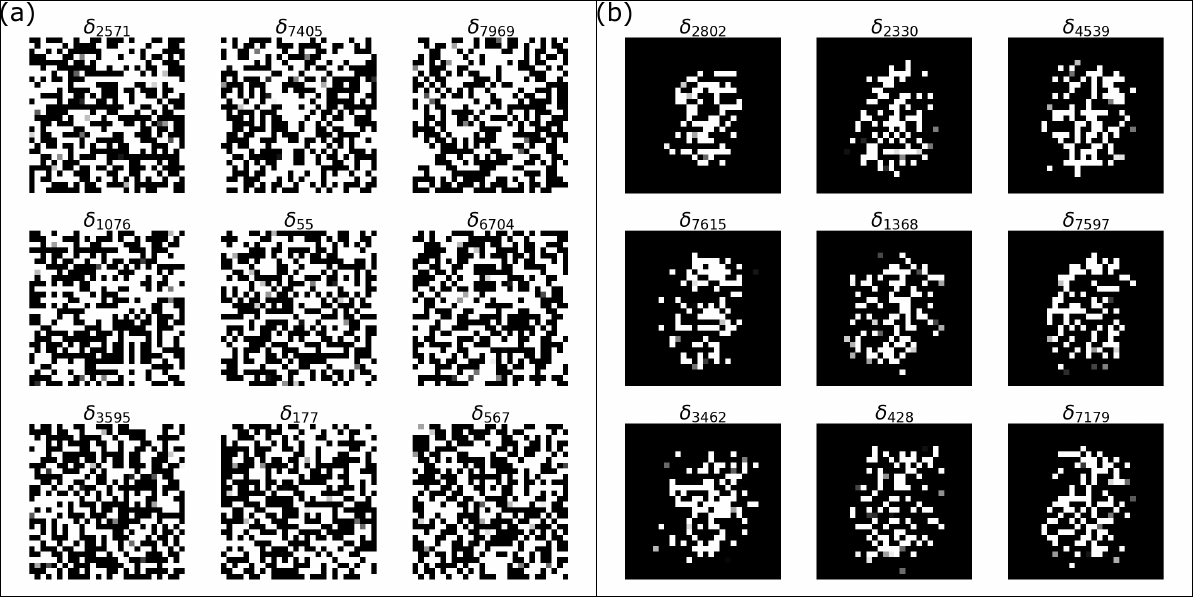}
		\caption{Visualization of features in the hidden layer on MNIST. For $j \in [2^{13}]$, $\left \|\vu_j \right \|_2$ is in descending order from left to right and top to bottom. (a) Without Diffusion. (b) With Diffusion. While the system without diffusion takes a random walk, the system with diffusion moves in the direction of real manifold of natural images in order to maximize excitation. }
		\label{mnist_sup_aug_feat_viz}
	\end{figure*}

\begin{figure}[!h]
		\centering
		\includegraphics[width=0.5\columnwidth]{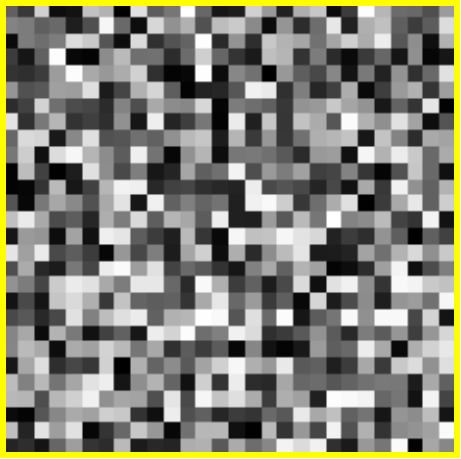}
		\caption{Input image used in the visualization of features.}
		\label{feat_viz_ip}
	\end{figure}

	As shown in Figure~\ref{mnist_sup_aug_feat_viz}, maximization of excitation without diffusion leads to perturbation in every spatial location of the input image. In contrast to that, the system with diffusion leads to perturbations in locations where essential information is present. From another perspective, this indicates excitation when the input images are pushed towards actual manifold of MNIST data. The fact that both start from the same initial point asserts this intriguing property of adversarial interaction. The input image is shown in Figure~\ref{feat_viz_ip}.
	
	\subsubsection{Results on FashionMNIST} 
\begin{figure}[!h]
		\centering
		\includegraphics[width=\linewidth]{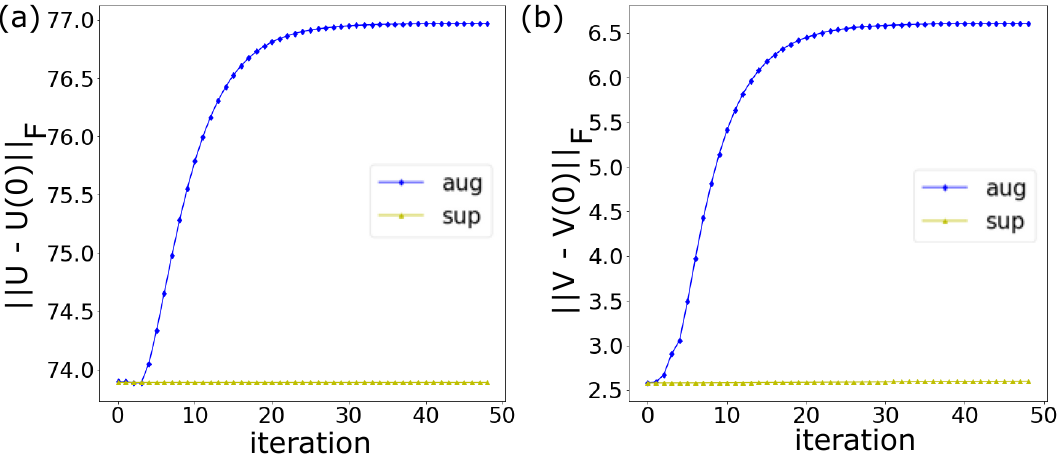}
		\caption{Distance from initialization in the (a) hidden layer (b) top layer on FashionMNIST.}
		\label{fmnist_sup_aug_part_dis_init}
	\end{figure}

As shown in Figure~\ref{fmnist_sup_aug_part_dis_init}, the distance from initialization is larger than synthetic datasets in both supervised and adversarial learning. One reason for such large deviation is the increased complexity on FashionMNIST. Moreover, it follows exponential trend in augmented objective and explores a larger subspace around initialization as predicted by our theory.

\begin{figure*}[!h]
		\centering
		\includegraphics[width=\linewidth]{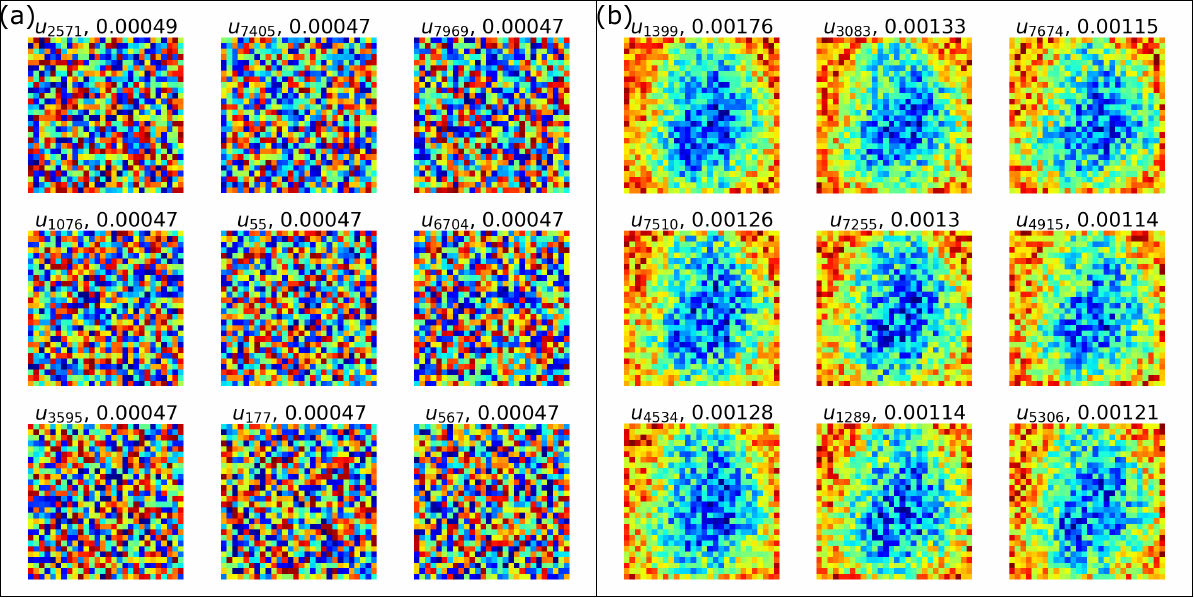}
		\caption{Comparing hidden layer filters on FashionMNIST. For $j \in [2^{13}]$, $\left \|\vu_j \right \|_2$ is in descending order from left to right and top to bottom. (a) Without Diffusion. (b) With Diffusion. High variance captures non-homogeneity in weight space. Adversarial interaction allows neurons in the hidden layer to focus on central region where useful information is accessible.}
		\label{fmnist_sup_aug_ker_var}
	\end{figure*}

	\textbf{Weight Analysis.} Similar to MNIST, we observe the self-organization tendency even on FashionMNIST. As shown in Figure~\ref{fmnist_sup_aug_ker_var}, local interaction with the advent of diffusion allows neurons in the hidden layer to cooperate while operating on the input samples.
	
\begin{figure*}[!h]
		\centering
		\includegraphics[width=\linewidth]{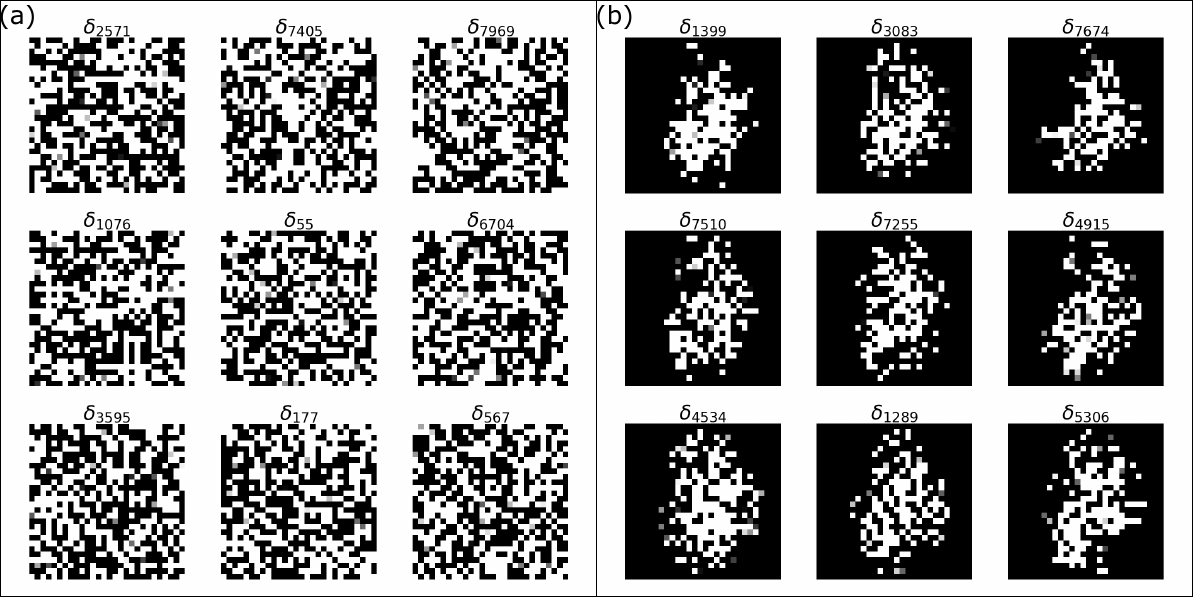}
		\caption{Visualization of features in the hidden layer on FashionMNIST. For $j \in [2^{13}]$, $\left \|\vu_j \right \|_2$ is in descending order from left to right and top to bottom. (a) Without Diffusion. (b) With Diffusion. The system with diffusion gets excited the most when input images are pushed towards actual manifold of real images. }
		\label{fmnist_sup_aug_feat_viz}
	\end{figure*}

	\textbf{Feature Visualization.} Figure~\ref{fmnist_sup_aug_feat_viz} provides further evidence of non-homogeneity in learned features. Recall that the task here is not to generate realistic looking images, but classify them into different categories. The discriminator does not have access to real images in this setting. It only sees the predicted and true labels in $\mathbb{R}^{10}$. Interestingly, adversarial interaction allows to learn the manifold of actual data while performing the desired classification task. Contrary to that, supervised learning seems to take a random walk in the feature space, suggesting misunderstanding of the true characteristics of natural images. 
	
	\subsection{Dissection of Diffusion} 
\begin{figure}[t]
		\centering
		\includegraphics[width=\linewidth]{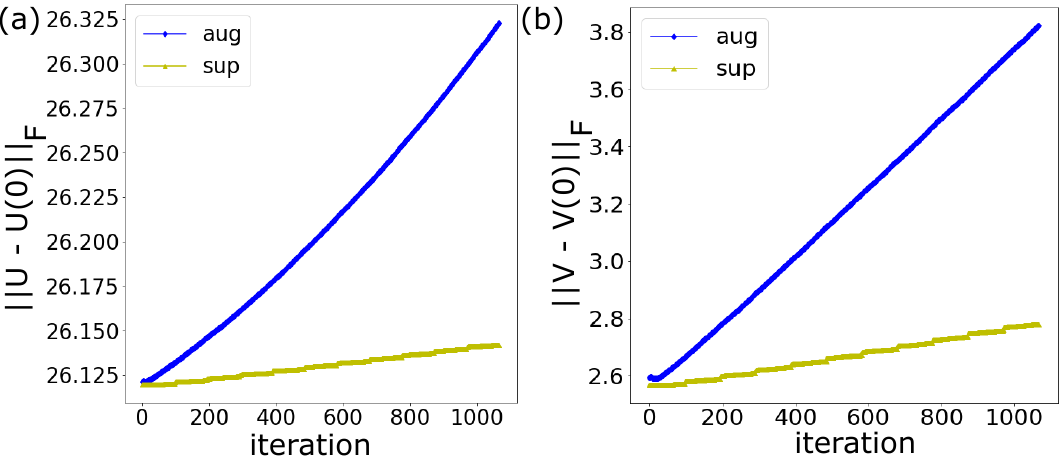}
		\caption{Distance from initialization in the (a) hidden layer (b) top layer on CIFAR10 with $2^{10}$ hidden units.}
		\label{cifar_sup_aug_part_dis_init}
	\end{figure}

\begin{figure}[t]
		\centering
		\includegraphics[width=\linewidth]{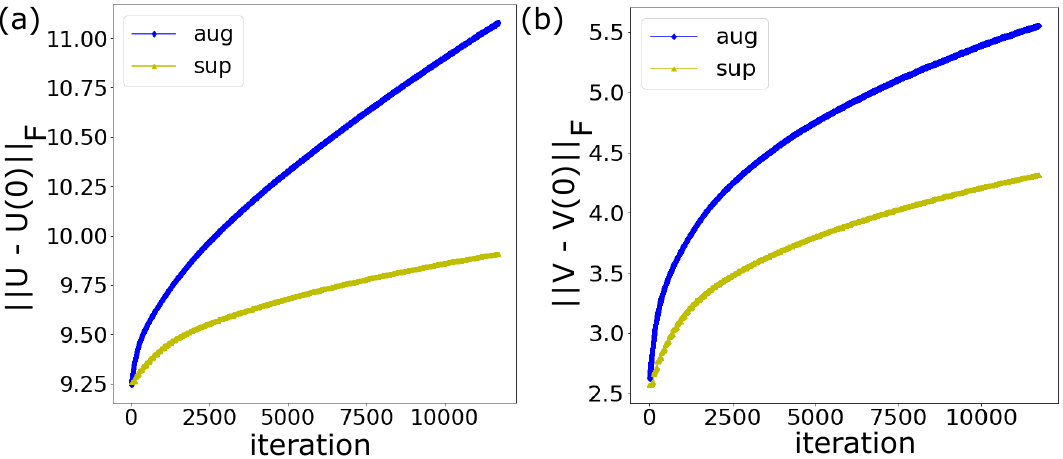}
		\caption{Distance from initialization in the (a) hidden layer (b) top layer on MNIST with $2^{7}$ hidden units.}
		\label{mnists_sup_aug_part_dis_init}
	\end{figure}

	To understand the contribution of diffusion term as given by equation~(\ref{dju}), we experiment with less number of hidden units on MNIST and CIFAR10 datasets. Recall that \textbf{Theorem 2} suggests a larger linear slope with smaller number of hidden units. Since reaction term settles down exponentially fast, a larger linear slope is expected to dominate the distance from initialization beyond the settling point. This hypothesis is supported by the experimental results in Figure~\ref{cifar_sup_aug_part_dis_init}~and~\ref{mnists_sup_aug_part_dis_init} where the dominance of linear slope is discernible. To this end, we have verified the assumptions and theorems empirically on two synthetic datasets and three benchmark datasets. In the succeeding discussion, we put emphasis on pattern formation using Turing's RD model, the developed PRD model, and Gray-Scott model.

	\subsection{Turing Patterns by RD Model}
	\label{tprdm}
	In this section, we provide numerical simulations to better understand the specific dynamics in RD model. Turing's fundamental equations capture the dynamics of patterns as a reminiscent of those seen in nature~\cite{turing1952chemical}:
	\begin{equation}
	\begin{split}
	\frac{\partial u}{\partial t} &= a(u-h) + b (v-k)+{\mu}'\nabla^2u\\
	\frac{\partial v}{\partial t} &= c(u-h) + d (v-k)+{\nu}'\nabla^2v.
	\end{split}
	\end{equation}
	Here, $u$ and $v$ denote two morphogens which attain equilibrium at $h$ and $k$ respectively. The reaction term is controlled by $a,b,c$ and $d$. On the other hand, the diffusion term is controlled by ${\mu}'$ and ${\nu}'$. 
	\subsubsection{Implementation Details}
\begin{figure}[!t]
		\centering
		\includegraphics[width=\linewidth]{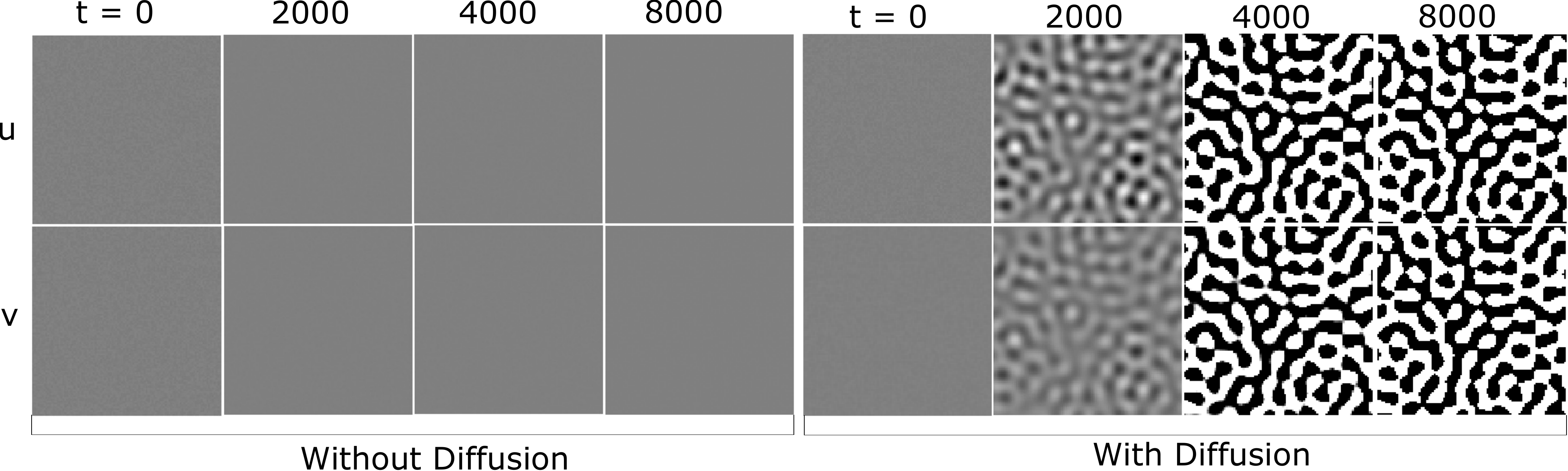}
		\caption{Turing pattern formation. The diffusible factors help break the symmetry and homogeneity.}
		\label{turing}
	\end{figure}

	Figure~\ref{turing} illustrates Turing patterns with parameters, $a=1,b=-1,c=3,d=-1.5,h=1,k=1,{\mu}'=0.0001, {\nu}'=0.0006$. The grid size is $[100\times100]$ and temporal resolution, $0.02s$.\footnote{One can select another set of parameters to create a different pattern. The present discussion is least affected by the choice of these parameters. However, it is essential to draw insights from the fundamental equations.}   
	
	\subsubsection{Analysis on Simulated Dataset} Observe that by discarding diffusion terms, i.e., ${\mu}'=0$ and ${\nu}'=0$, the system attains homogenous equilibrium with $u_{eq}=h$ and $v_{eq}=k$. For this reason, the morphogens, $u$ and $v$ are initialized with $u_{eq}+\mathcal{U}(-0.03,0.03)$ and $v_{eq}+\mathcal{U}(-0.03,0.03)$. The uniformly distributed perturbation, $\mathcal{U}(.)$ helps break homogeneity and spatial symmetry when diffusible factors are introduced in the system. Figure~\ref{turing} supports this hypothesis as we observe emergence of evolutionary patterns as a result of diffusion. Notice that the diffusible factors are orders of magnitude smaller than reaction coefficients. This mild diffusion is enough to induce Turing instability that results in evolutionary patterns. 
	
	\subsection{Turing-Like Patterns by PRD Model}
	\label{tlprd}
	\subsubsection{Implementation Details} Here, Turing-like patterns contain $m$ neurons each in the hidden layer ($\vu$) and top layer ($\vv$). For visualization purpose, the neurons in $\mathbb{R}^{d_{in}}$ are mapped to $\mathbb{R}^{10}$ using Principal Component Analysis (PCA). Further, t-Distributed Stochastic Neighbor Embedding (t-SNE)~\cite{maaten2008visualizing} is employed to map each neuron to $\mathbb{R}^{2}$. t-SNE is particularly well-suited for visualizing high dimensional data. However, sometime it requires a first stage dimensionality reduction due to its high computational cost. In the top layer, t-SNE is directly employed to map from $\mathbb{R}^{10}$ to $\mathbb{R}^2$.  To reduce computational bottleneck, we have selected initial 2048 neurons out of $2^{13}$ in the hidden layer for visualization. The absolute scales of these plots are irrelevant to the present body of analysis. In each pattern, different colors are used to represent different clusters of neurons for the sole purpose of visual assimilation. For clustering purpose, Affinity Propagation (AP) is used to capture the essence of node connectivity during information propagation among hidden units~\cite{frey2006mixture,frey2007clustering}.
	
	\subsubsection{Analysis on Synthetic Datasets}
\begin{figure}[!t]
		\centering
		\includegraphics[width=\linewidth]{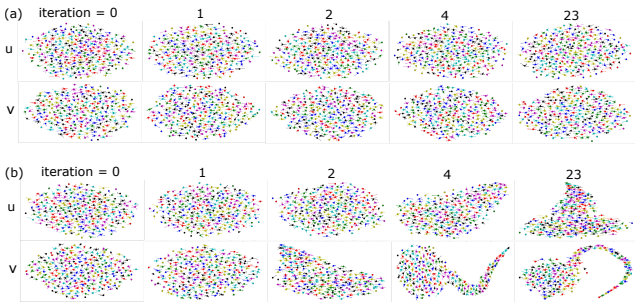}
		\caption{Pattern formation on synthetic data, $d_{in}=784$. (a) Without Diffusion. (b) With Diffusion.}
		\label{syn1_sup_aug}
	\end{figure}

\begin{figure}[!t]
		\centering
		\includegraphics[width=\linewidth]{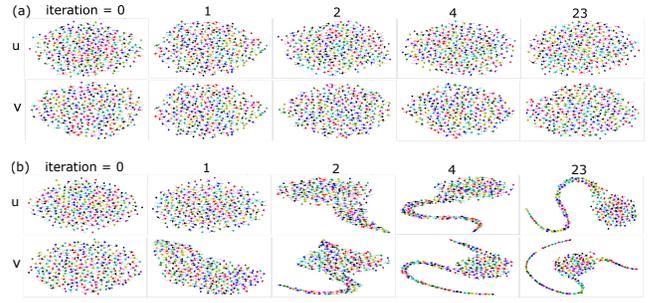}
		\caption{Pattern formation on synthetic data, $d_{in}=1024$. (a) Without Diffusion. (b) With Diffusion.}
		\label{syn2_sup_aug}
	\end{figure}

	Similar to Turing's RD model, we observe spatially symmetric and homogeneous equilibrium in the absence of adversarial regularization. The system retains symmetry over iterations as long as it does not receive any signal that may induce diffusibility. As shown in Figure~\ref{syn1_sup_aug}(a) and  Figure~\ref{syn2_sup_aug}(a), the neurons in the hidden layer ($u$) and top layer ($v$) lie close to their initial values. While gradient descent finds an $\epsilon$-stationary point, the neurons in both layers do not deviate much from their initial topology. Consequently, the spatial symmetry and homogeneity is preserved throughout training. 
	
	On the other hand, with the introduction of an adversary into the system, the exchange of signal between generator and discriminator constitutes the basis of diffusible factors. Evident from Figure~\ref{syn1_sup_aug}(b) and Figure~\ref{syn2_sup_aug}(b), the adversarial interaction helps break the inherent spatial symmetry and homogeneity while randomly initialized gradient descent still finds $\epsilon$-stationary point.  In supervised and adversarial setting, the qualitative observation of Turing-like patterns favors the quantitative analyses made in \textbf{Experimental Results}. Furthermore, \textbf{Theorem 3} provides a justification to late breakdown of symmetry in the hidden layer compared to top layer. Since $\mathfrak{D}^{\vu}_{j}\left ( \nabla^2 \vu_j \right )$ has a tighter asymptotic bound compared to $\mathfrak{D}^{\vv}_{j}\left ( \nabla^2 \vv_j \right )$, it has a lower rate of diffusion on the experimented datasets.
	
	\subsubsection{Analysis on MNIST and FashionMNIST} 
\begin{figure}[!h]
		\centering
		\includegraphics[width=\linewidth]{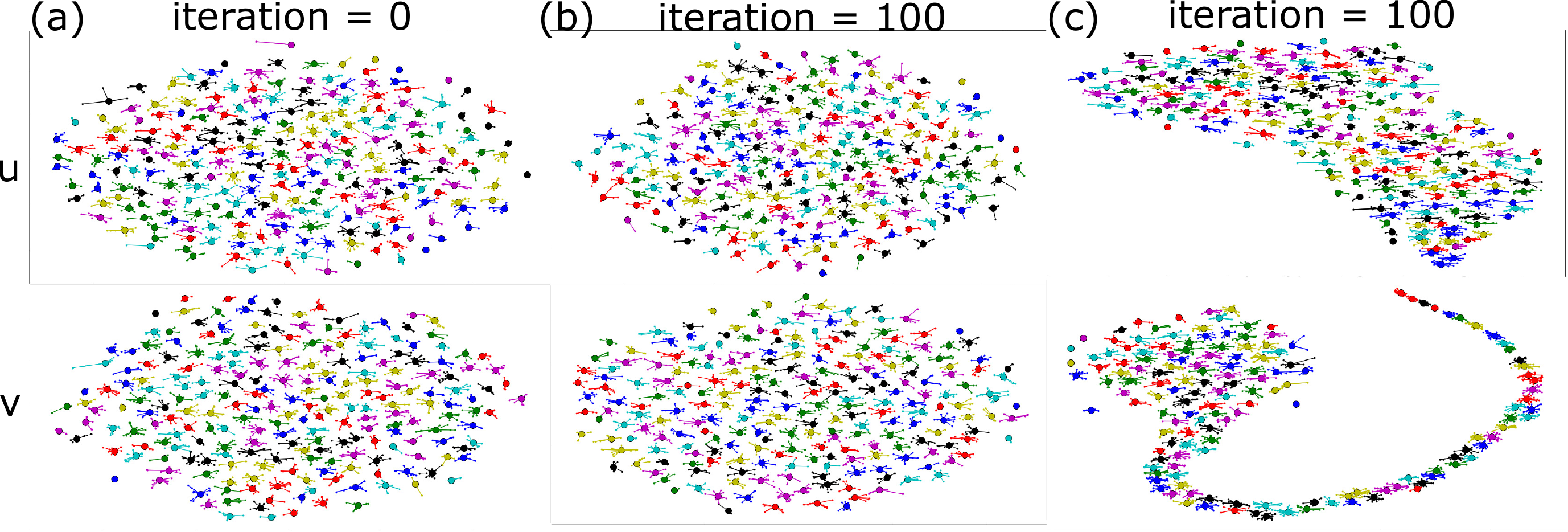}
		\caption{Pattern formation on MNIST. (a) Initialization. (b) Without Diffusion. (c) With Diffusion.}
		\label{mnist_rd}
	\end{figure}

\begin{figure}[!h]
		\centering
		\includegraphics[width=\linewidth]{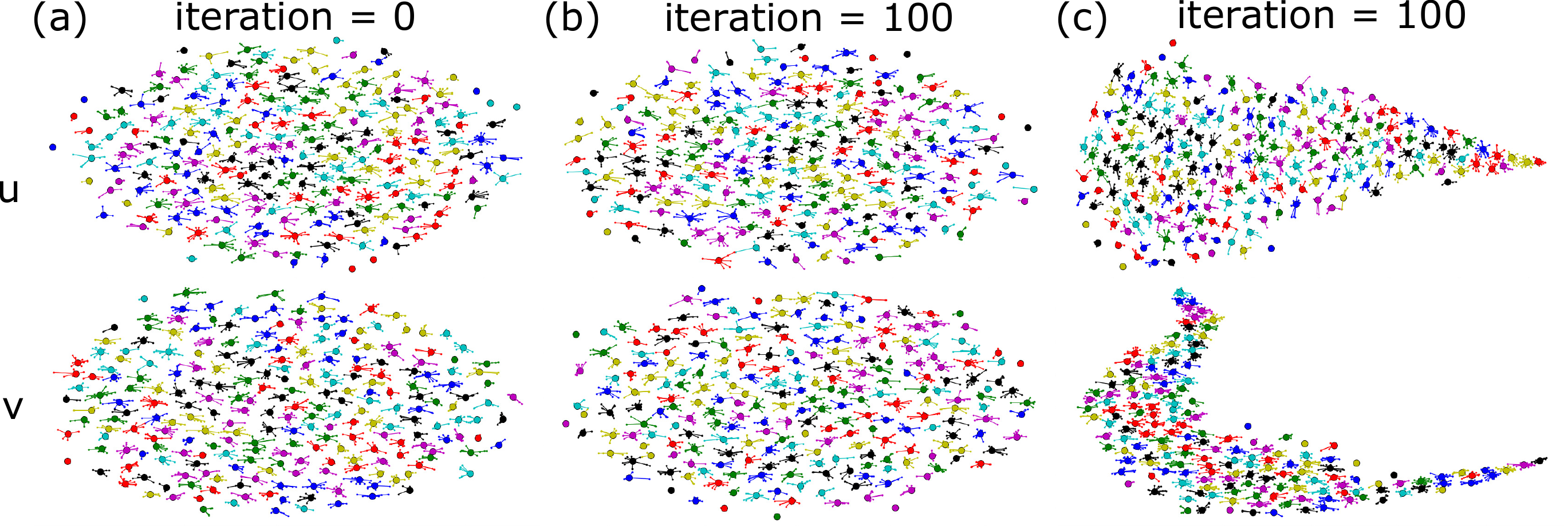}
		\caption{Pattern formation on FashionMNIST. (a) Initialization. (b) Without Diffusion. (c) With Diffusion.}
		\label{fmnist_rd}
	\end{figure}

\begin{figure}[!h]
		\centering
		\includegraphics[width=\linewidth]{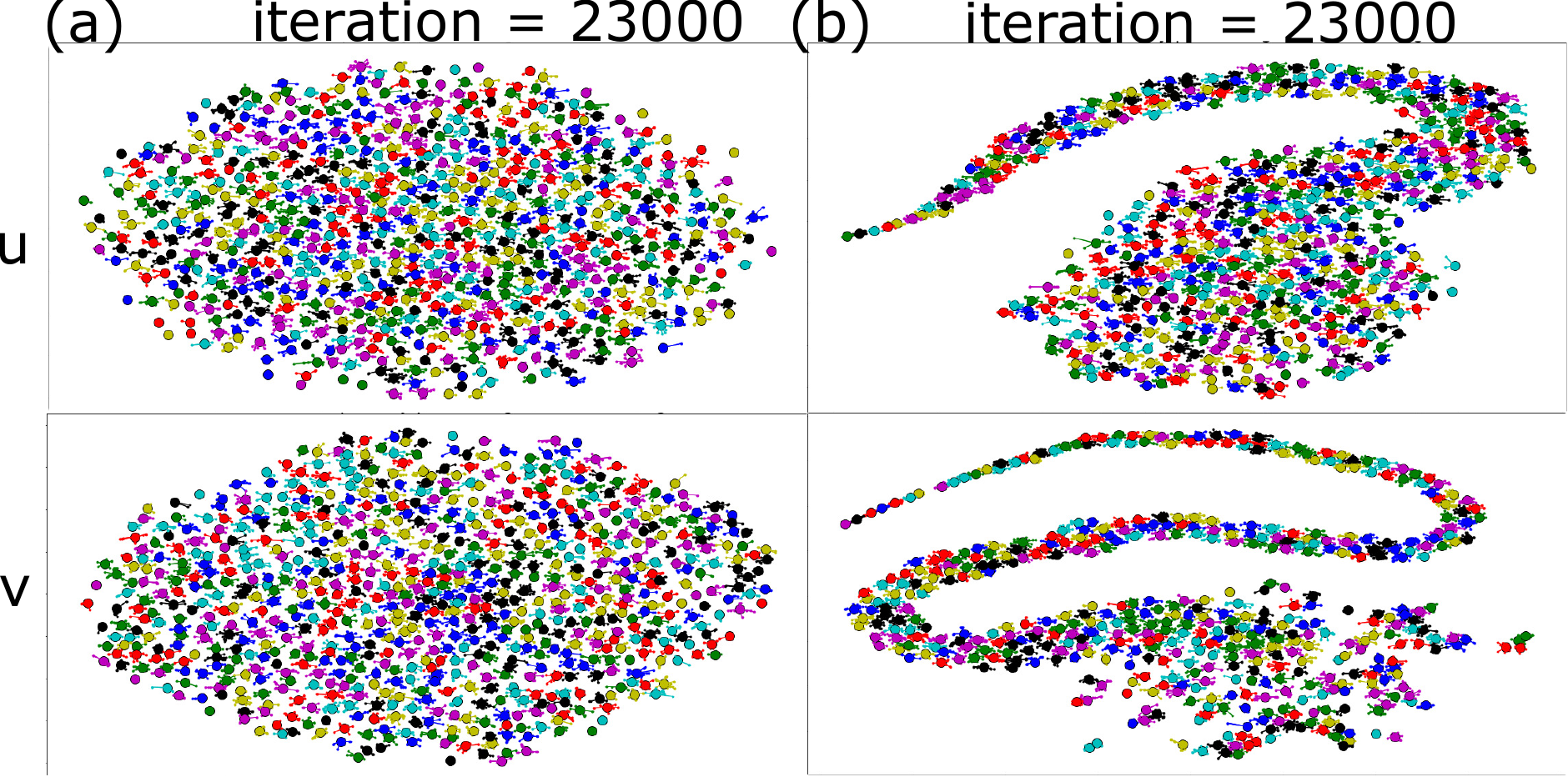}
		\caption{Pattern formation on MNIST. Visualization of all $2^{13}$ neurons. (a) Without Diffusion. (b) With Diffusion.}
		\label{mnist_rd_full}
	\end{figure}

Interestingly, the simulated patterns resemble those observed in the hidden layer and top layer when trained on actual datasets: MNIST (Figure~\ref{mnist_rd}) and FashionMNIST (Figure~\ref{fmnist_rd}). Here, similarity is measured in a topological sense that reflects breakdown of symmetry and homogeneity under adversarial interaction.  	Figure~\ref{mnist_rd_full} illustrates Turing-like patterns formed by all neurons after fully trained on MNIST.	
	
	\subsection{Turing-Like Patterns by Gray-Scott Model}
	\label{tlpgs}
	For better understanding of RD systems, we have also simulated Gray-Scott patterns~\cite{gray1984autocatalytic}\footnote{\cite{sayama2015introduction} provides a comprehensive overview of Gray-Scott pattern formation subject to different parameters.}:
	\begin{equation}
	\begin{split}
	\frac{\partial u}{\partial t} &= F\left ( 1-u \right ) - uv^2 +{\mu }'\nabla^2u\\
	\frac{\partial v}{\partial t} &= -\left ( F+k \right )v + uv^2 +{\nu }'\nabla^2v
	\end{split}
	\end{equation}
	\subsubsection{Implementation Details}
\begin{figure}[!h]
		\centering
		\includegraphics[width=\linewidth]{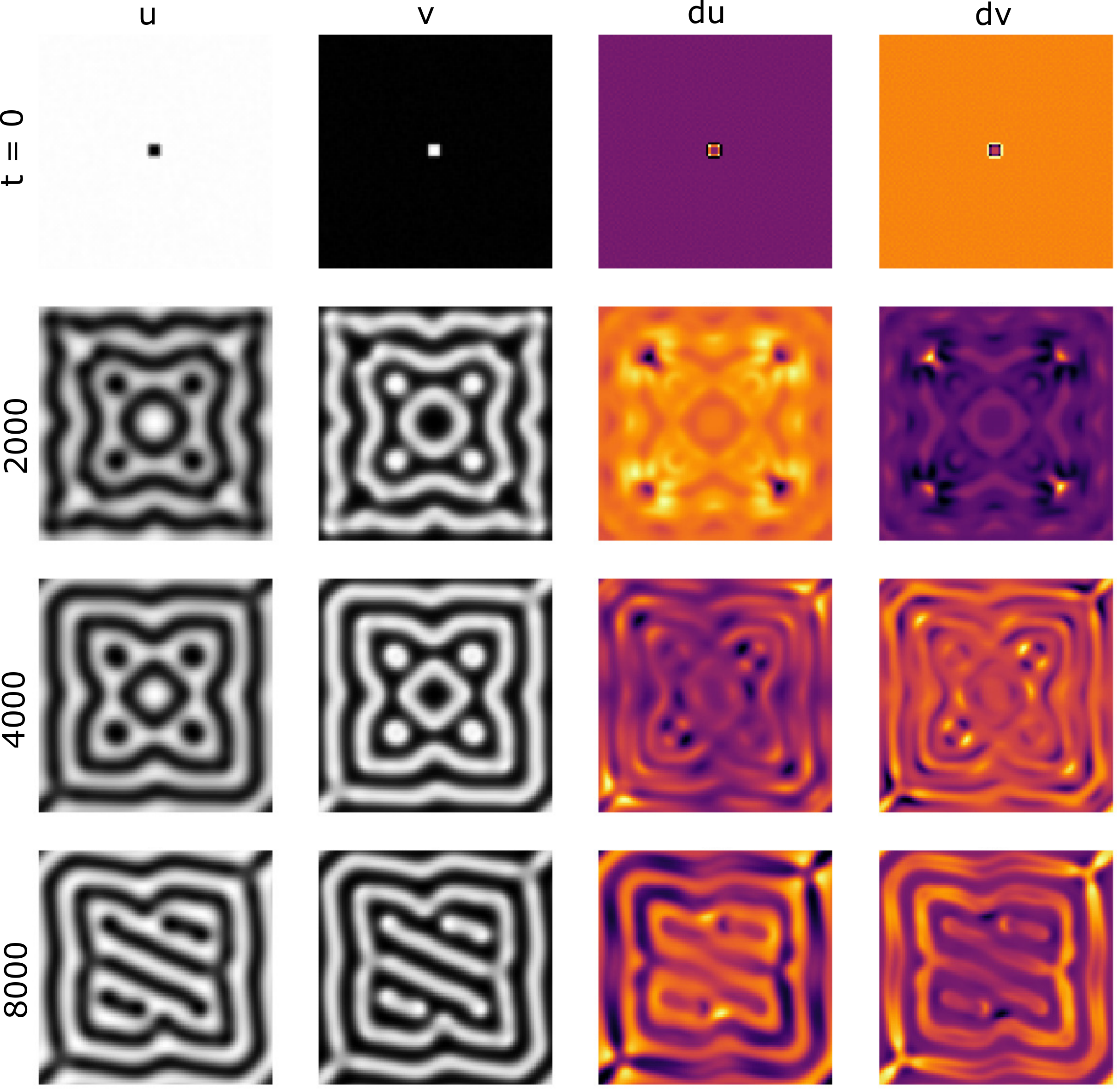}
		\caption{Gray-Scott pattern formation. $F=0.025$ and $k=0.055$.}
		\label{gs1}
	\end{figure}

\begin{figure}[!h]
		\centering
		\includegraphics[width=\linewidth]{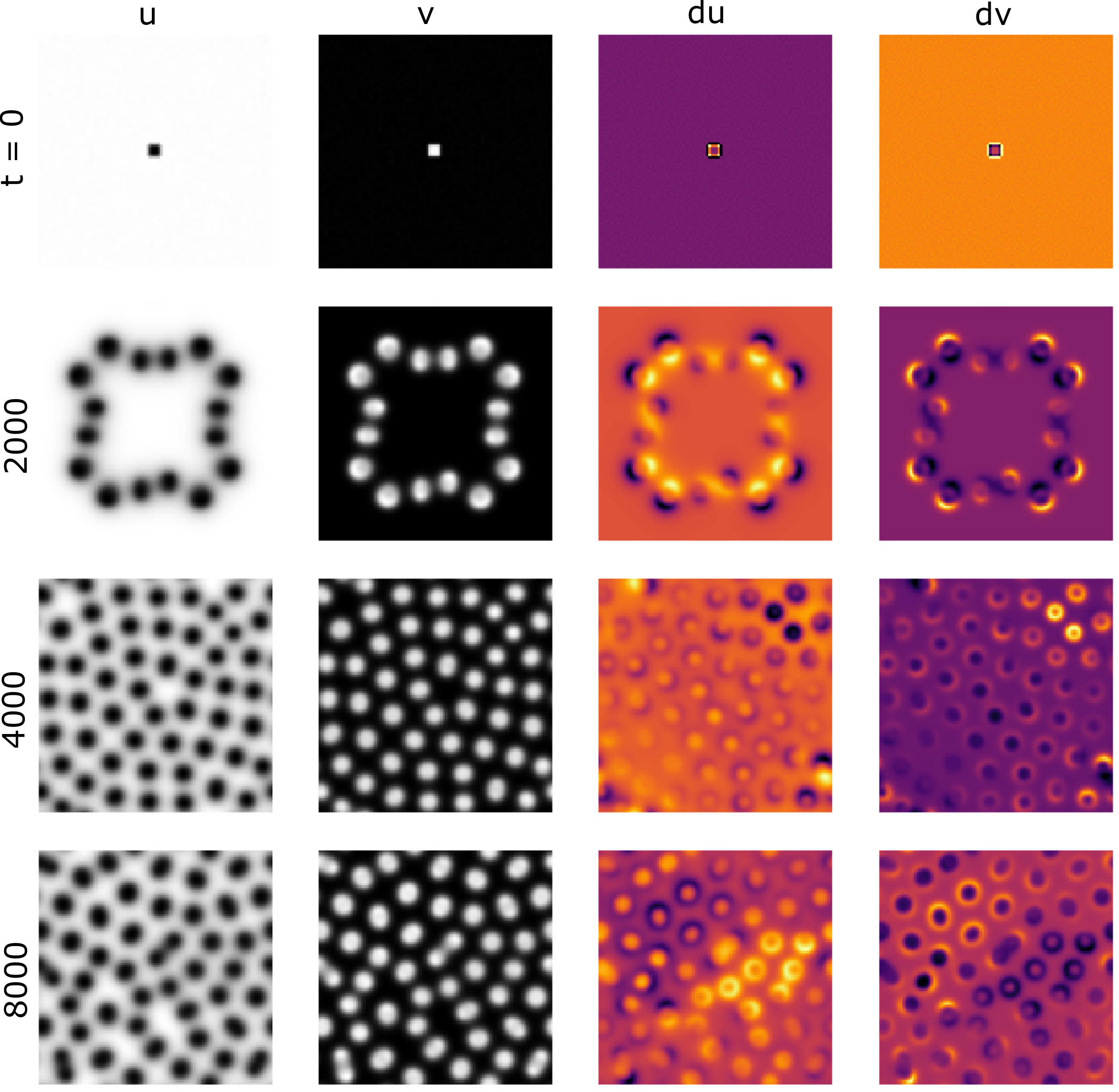}
		\caption{Gray-Scott pattern formation.  $F=0.025$ and $k=0.060$.}
		\label{gs2}
	\end{figure}

\begin{figure}[!h]
		\centering
		\includegraphics[width=\linewidth]{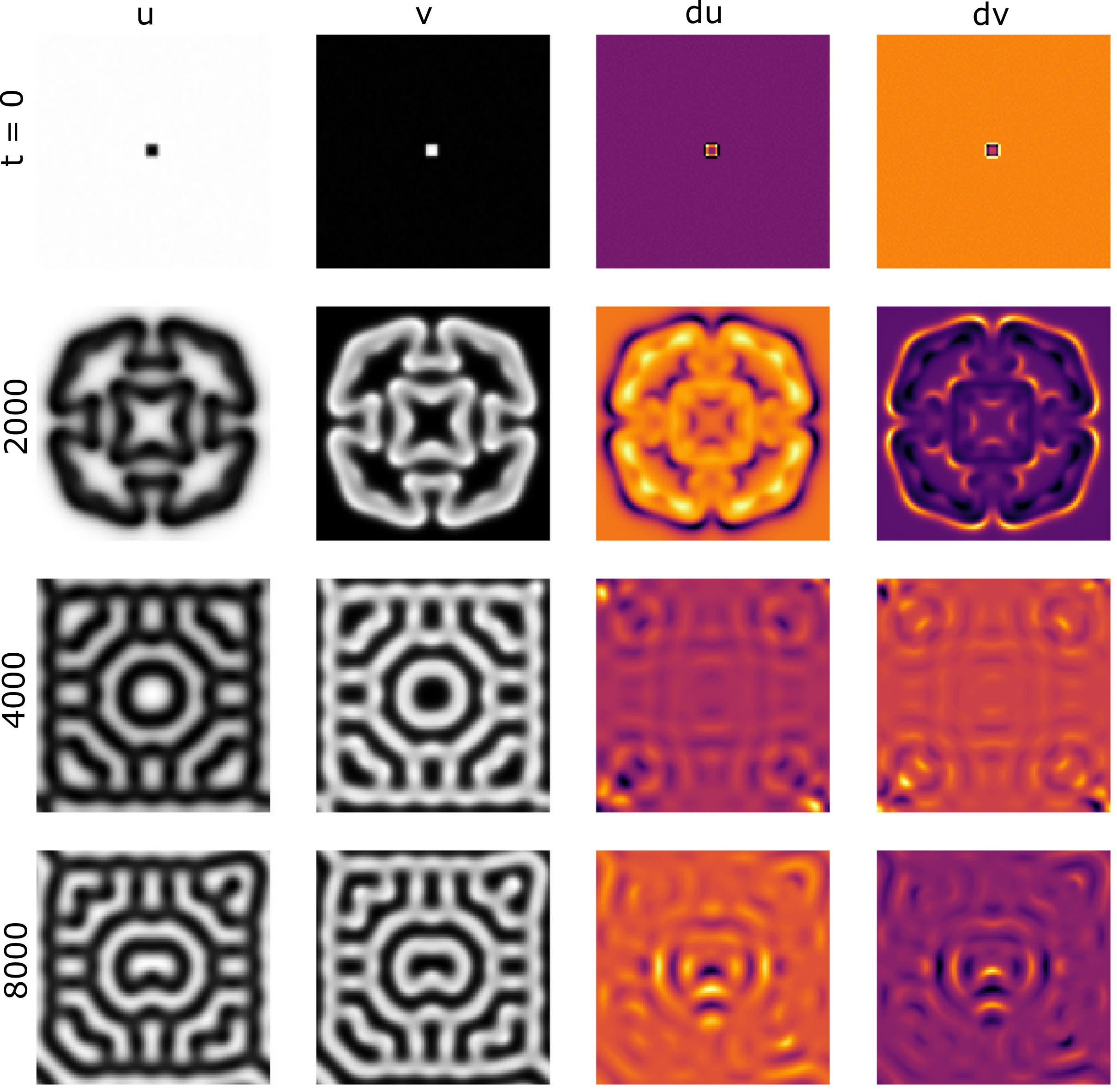}
		\caption{Gray-Scott pattern formation.  $F=0.040$ and $k=0.060$.}
		\label{gs3}
	\end{figure}

\begin{figure}[!h]
		\centering
		\includegraphics[width=\linewidth]{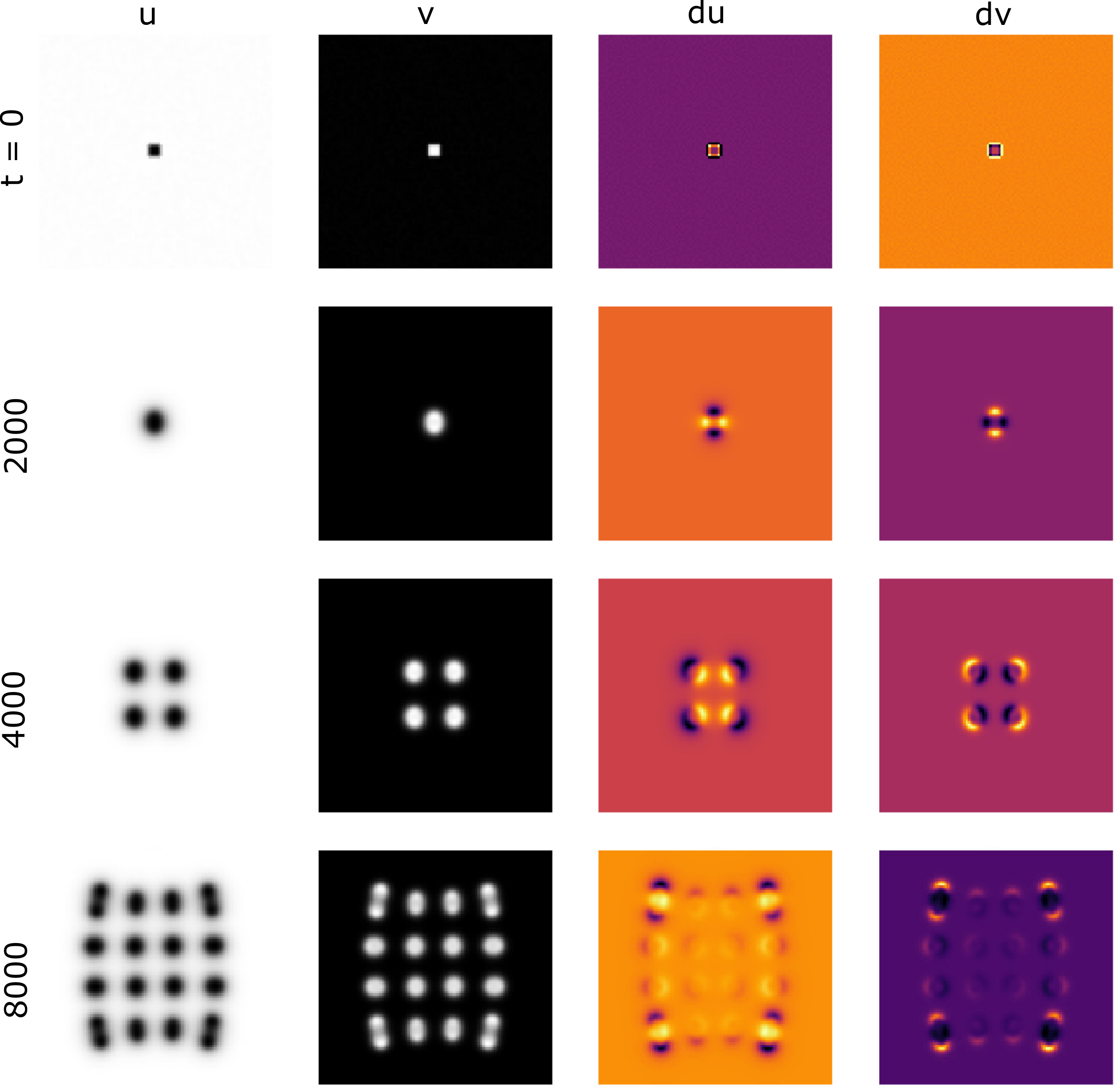}
		\caption{Gray-Scott pattern formation.  $F=0.035$ and $k=0.065$.}
		\label{gs4}
	\end{figure}

	The morphogens lie on a two dimensional grid of size $[100\times100]$. The initial conditions are such that there exist certain concentration difference in each morphogen which gives rise to diffusion. The grid therefore is initialized by all ones ($u$) and all zeros ($v$) with corresponding central $[5\times5]$ patch inverted as shown in Figure~\ref{gs1}, \ref{gs2}, \ref{gs3} and \ref{gs4} at $t=0$. With very high rate of diffusion, the morphogens get destroyed too quickly to have a positive influence on the reactions, which are the most important parts of the assumptions in RD model. Therefore, the diffusion coefficients are set to ${\mu}' = 2e-5$ and ${\nu}' = 1e-5$ in the present analysis of Gray-Scott patterns. 
	
	\subsubsection{Analysis on Simulated Dataset} By changing the values of $F$ and $k$, a wide variety of patterns can be created. In all these figures, different irregularities induce different patterns despite exactly identical intitial condition. This is a crucial observation as it justifies non-stationary Turing-like patterns in the intermediate iterations of adversarial training as shown in Figure~\ref{syn1_sup_aug}(b) and Figure~\ref{syn2_sup_aug}(b). In this case, different irregularities get introduced at each iteration due to stochasticity in the finite sum optimization problem.
	
\end{document}